\documentclass{article}

\usepackage{PRIMEarxiv}

\usepackage[utf8]{inputenc} 
\usepackage[T1]{fontenc}    
\usepackage{hyperref}       
\usepackage{url}            
\usepackage{booktabs}       
\usepackage{amsfonts}       
\usepackage{nicefrac}       
\usepackage{microtype}      
\usepackage{graphicx}
\usepackage{booktabs}
\usepackage{lscape}

\usepackage{amsmath}
\usepackage{lscape}

\newcommand{\paratitle}[1]{\vspace{1.5ex}\noindent \textbf{#1}}

\usepackage[utf8]{inputenc}
\usepackage{array}
\usepackage[normalem]{ulem}
\usepackage{lscape}
\usepackage[paper=portrait,pagesize]{typearea}
\usepackage{placeins}
\usepackage{longtable}
\usepackage{tabu}
\usepackage{float}
\useunder{\uline}{\ul}{}
\usepackage{multirow}
\usepackage[table,xcdraw]{xcolor}
\usepackage{graphicx}
\usepackage{subfig}
\usepackage[english]{babel}
\usepackage{geometry}
\usepackage{tikz}
\usetikzlibrary{mindmap}
\usepackage[backend=biber,style=numeric,sorting=ynt]{biblatex}
\usepackage{pifont}

\title{Deep Learning meets Liveness Detection: Recent Advancements and Challenges

}

\author{
  Arian Sabaghi\thanks{Corresponding Author}\\
  Amirkabir University of Technology(Tehran Polytechnic), Tehran, Iran\\
  \texttt{arian.s@aut.ac.ir} \\
\And
  Marzieh Oghbaie\\
  Imam Khomeini International University, Qazvin, Iran \\
  \texttt{marzieh.oghbaie@edu.ikiu.ac.ir} \\
\And
  Kooshan Hashemifard  \\
  University of Alicante, Alicante, Spain\\
  \texttt{k.hashemifard@ua.es} \\
\And
  Mohammad Akbari  \\
  Amirkabir University of Technology(Tehran Polytechnic), Tehran, Iran\\
  School of Computer Science, Institute for Research in Fundamental Sciences (IPM), Tehran, Iran\\
  \texttt{akbari.ma@aut.ac.ir} \\

}

\begin{document}
\maketitle

\begin{abstract}
Facial biometrics has been recently received tremendous attention as a convenient replacement for traditional authentication systems. Consequently, detecting malicious attempts has found great significance, leading to extensive studies in face anti-spoofing~(FAS),i.e., face presentation attack detection.
Deep feature learning and techniques, as opposed to hand-crafted features, have promised dramatic increase in the FAS systems' accuracy, tackling the key challenges of materializing real-world application of such systems. 
Hence, a new research area dealing with development of more generalized as well as accurate models is increasingly attracting the attention of the research community and industry. In this paper, we present a comprehensive survey on the literature related to deep-feature-based FAS methods since 2017. To shed light on this topic, a semantic taxonomy based on various features and learning methodologies is represented. Further, we cover predominant public datasets for FAS in a chronological order, their evolutional progress, and the evaluation criteria (both intra-dataset and inter-dataset). Finally, we discuss the open research challenges and future directions.
\end{abstract}


\section{Introduction}
\label{sec:background}
With the advancement of face recognition systems~\cites{deng2019arcface, wang2018cosface, cao2018vggface2, schroff2015facenet, taigman2014deepface} in surpassing humans' accuracy, many domains such as law enforcement~\cite{lynch2020face} and retail industries~\cites{ntech, sightcorp, facefirst} have leveraged such systems in their downstream applications. Face recognition systems bring convenience and much-needed security for users and reliability for business owners. For instance, face recognition can be applied in shopping centers to detect VIP or blacklisted patrons at high speed and accuracy or let consumers easily carry out tasks like opening a bank account, which previously had demanded their physical presence. 
Despite their advantages, face recognition systems are threatened by impersonation attacks~\cites{liu2019deep}~(impersonating other people by providing their image) and obfuscation attacks~\cites{liu2019deep}~(hiding true identity). As a result, Presentation Attack Detection~(PAD) or Face Anti-Spoofing~(FAS) has received tremendous attention in recent years to validate the liveness of a given face image.
In some industrial use cases, such as accessing secure buildings, utilizing special hardware or sensors such as Time of Flight~(ToF), 3D Structured Light~(SL), Near-InfraRed~(NIR), thermal sensors, etc., simplifies differentiation between live faces and spoof attacks. Although these specific sensors offer an acceptable level of performance and accuracy, deploying them as a common solution is restricted. Properly speaking, the mostly available capturing devices are commercial RGB cameras lacking any specific sensor.
A basic alternative to compensate for the lack of specialized hardware is to rely on involuntary responses, such as eye blinking, smiling, or to ask user to perform a specific action, or series of actions. However, this approach is not user-friendly and cannot be used in applications where users’ cooperation is not plausible (surveillance camera). More importantly, users may attack them by replaying a video of a subject in front of the camera~(known as video replay attacks).


A wide range of FAS methods and techniques have been employed in both industrial and academic works to improve the accuracy and reduce the error rate. The very first attempts in FAS systems were based on hand-crafted features extracted from raw face images. In such traditional approaches, hand-crafted features, such as Local Binary Pattern~(LBP)~\cite{ojala1994performance}, Local Phase Quantization~(LPQ)~\cite{ojansivu2008blur}, Histogram of Oriented Gradients~(HOG)~\cite{freeman1995orientation}, are extracted and then fed to a machine learning classifier like Support Vector Machines~(SVM)~\cite{cristianini2000introduction} or a fully-connected layer for final classification. Such approaches are not limited by the mentioned barriers, although they only extract low-level features lacking enough discrimination.

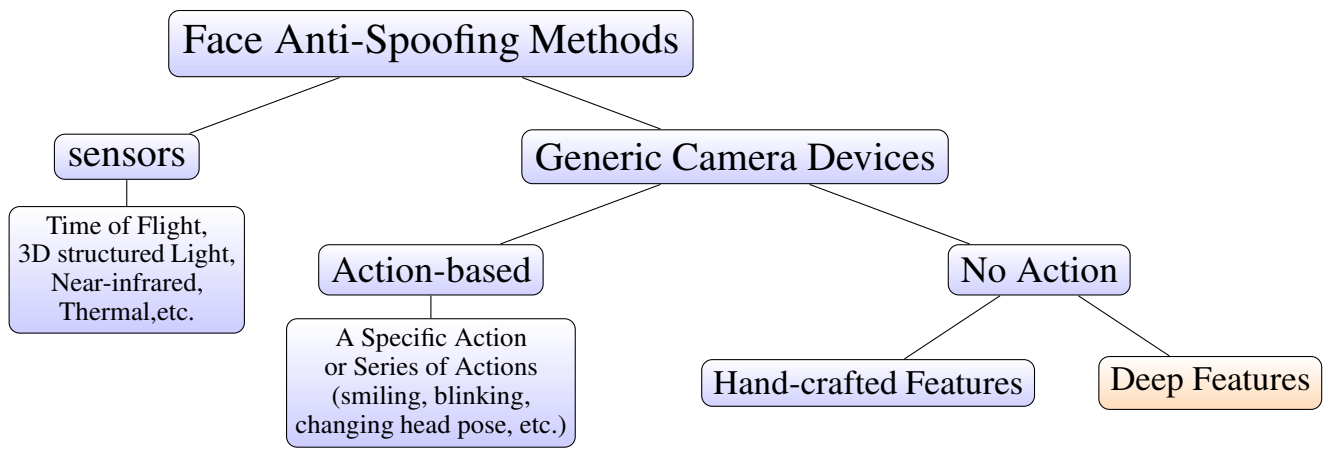
\begin{figure*}[!ht] 
\centering
\begin{tikzpicture}[sibling distance=10em,
  every node/.style = {shape=rectangle, rounded corners,
    draw, align=center,
    top color=white, bottom color=blue!20,},
    level 1/.style = {sibling distance=8cm},
    level 2/.style = {sibling distance=8cm},
    level 3/.style = {sibling distance=4.5cm}]
  \node [scale=1.6]{Face Anti-Spoofing Methods}
    child { node [scale=1.5]{sensors} 
       child {node {Time of Flight,\\3D structured Light,\\Near-infrared,\\Thermal,etc.} }
       }
    child { node [scale=1.5]{Generic Camera Devices}
       child { node [scale=1.4]{Action-based}
         child { node {A Specific Action\\or Series of Actions\\ (smiling, blinking,\\changing head pose, etc.)}}
       }
       child { node [scale=1.4]{No Action}
         child { node [scale=1.3]{Hand-crafted Features}}
         child { node [scale=1.3][bottom color=orange!30]{Deep Features}}
       }
    };
\end{tikzpicture}
\caption{\label{fig:tree} Face Anti-spoofing Methods.}
\end{figure*}

The popular success of deep learning various computer vision tasks results into employing deep neural models for FAS, which not only achieve superior performance over traditional ones but also are recognized as the dominant and most prevalent approach in FAS systems. 
Such systems process the input face using Convolutional Neural Networks~(CNN) to extract visual features to detect spoof attacks. Although these methods have reached acceptable performance in intra-dataset evaluations, their decisions on unseen data are not reliable. This is owing to the fact that simply using a binary classifier on top of CNN layers is prune to capture any distinguishing feature on training datasets~(e.g., screen bezel) instead of liveness related ones.
In this paper, we exclusively focus on methods in which deep learning plays the lead role because of their superiority and the fact that they do not need any specific video recording device.
Since a congregated overview and insightful categorization of these methods can be beneficial, in this paper, we try to understand the general spectrum of state-of-the-art FAS approaches and give a broad taxonomy of FAS methods (figure~\ref{fig:mindmap}), as well as a brief description for each one. Moreover, a high-level taxonomy is given (figure~\ref{fig:tree}) to clarify FAS methods based on deep features in relation to other approaches. Note that this survey neither cover methods deploying a specific action nor those merely based on hand-crafted features. 


\subsection{Our Contributions}

Before dive into details, it is worth noting that this paper only considers application scenarios wherein capturing devices are commonly available for daily usage and overlooks devices that rely on specific hardware not available commercial RGB camera's of users devices~(e.g. tablets, smart phones, etc.).

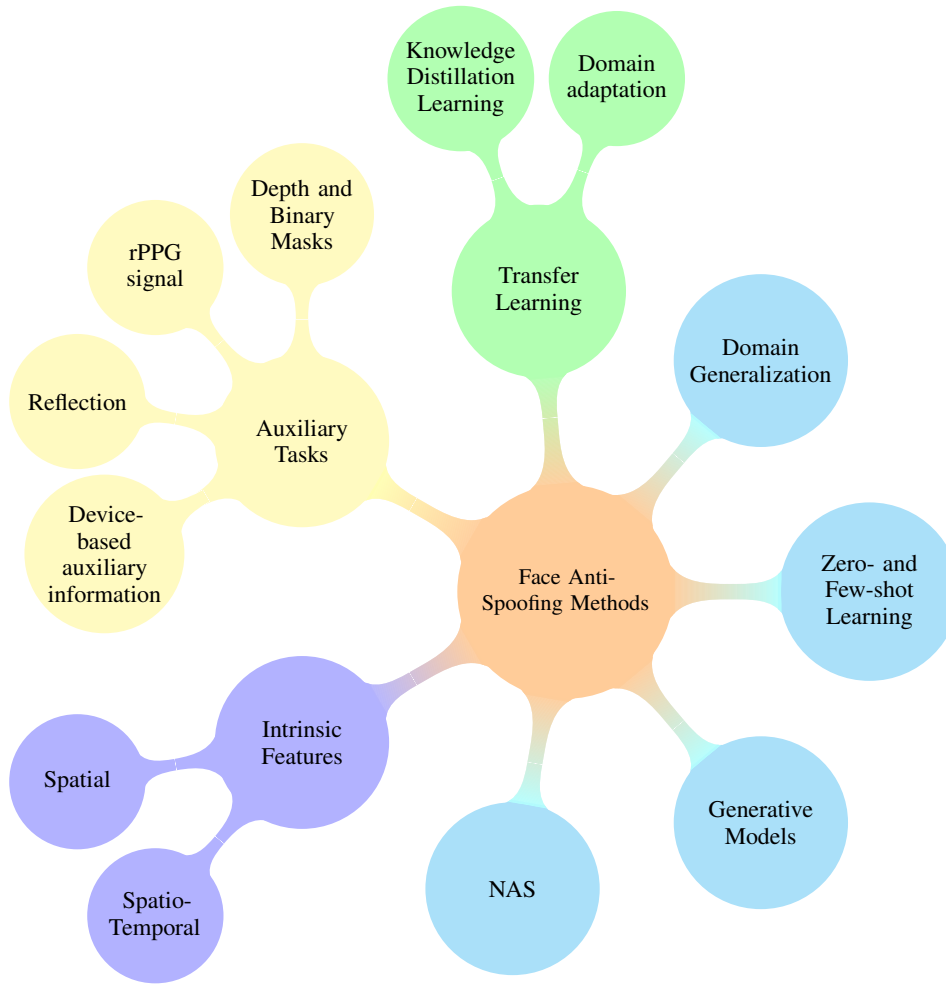
\begin{figure*}[h!]
\centering
\begin{tikzpicture}[mindmap, grow cyclic, every node/.style=concept, concept color=orange!40, 
    level 1/.append style={level distance=4cm,sibling angle=50},
    level 2/.append style={level distance=3cm,sibling angle=40},
    level 3/.append style={level distance=3cm,sibling angle=40}]

\node [ scale=0.7] {Face Anti-Spoofing Methods}
    child [concept color=blue!30] { node {Intrinsic Features}
        child { node {Spatial}}
        child { node {Spatio-Temporal}}
        }
        child [concept color=cyan!30] { node {NAS}}
        child [concept color=cyan!30] { node {Generative Models}}
        child [concept color=cyan!30] { node {Zero- and Few-shot Learning}}
        child [concept color=cyan!30] { node {Domain Generalization}}
    child [concept color=green!30, grow=95, level distance=4cm] { node {Transfer Learning}
        child [grow=70, level distance=3cm] { node  {Domain adaptation}}
        child [grow=110, level distance=3cm] { node  {Knowledge Distillation Learning}}}
    child [concept color=yellow!30] { node {Auxiliary Tasks}
        child { node {Depth and Binary Masks}}
        child { node {rPPG signal}}
        child {node  {Reflection}}
        child {node  {Device-based\\*auxiliary\\*information}}
    };

\end{tikzpicture}
\caption{\label{fig:mindmap}Face Anti-spoofing Methods.}
\end{figure*}

The most recent surveys on FAS methods are conducted by Yu et al.~\cite{yu2021deep} and Ming et al.~\cite{ming2020survey}, comprising a great number of studies and datasets. On one hand,~\cite{yu2021deep} covers numerous studies and datasets but it does not provide adequate description about the research questions and the corresponding solutions.
On the other hand,~\cite{ming2020survey} does not include a great number of influential works and datasets published recently.  We narrow down our study to the most recent works (published in 2017 onward) that use deep neural networks and mainly focus on technical advancement. For better differentiation among a diverse range of recent methods, we have semantically categorized them as follows (Figure~\ref{fig:mindmap}):

\textbf{Auxiliary tasks.} Since handling face anti-spoofing problem as a binary classification fails to learn generelizable features, some researchers have exploited human-defined cues, a.k.a auxiliary tasks, to extract task-related features. Auxiliary tasks include facial depth map, rPPg signal that corresponds to skin color tone variations, reflection, and device-based tasks, such as reflection due to mobile phone's screen.

\textbf{Intrinsic features.} Most vision tasks are accomplished through extracting discriminative features via proper deep neural structures. In case of face anti-spoofing, a large number of research solely rely on spatial or spatio-temporal features of deep neural networks to detect spoof attacks. While these methods may contain auxiliary tasks such as depth, their main focus revolve around network structure.

\textbf{Transfer learning.} In transfer learning, knowledge of a different but related task is transferred to a new task at hand. However, in face anti-spoofing, knowledge of the same task from a richer domain is transferred to another domain with less data. We cover two subcategories of transfer learning, namely domain adaptation and knowledge distillation. Domain adaptation employs similarity metrics or adversarial learning to achieve generalized feature space between source and target domain, and in knowledge distillation a teacher-student framework is adopted to do so.

\textbf{Domain Generalization.} Domain generalization takes similar measures to domain adaptation to find a generalized feature space across multiple data domains. However, in domain generalization there is no access to target domain, and only source domains are leveraged to this end. It is worth noting that in both Domain generalization and adaptation typically each dataset is considered as a separate domain, but other differentiating factors such as acquisition device can be employed to define different domains. 

\textbf{Zero- and Few-shot Learning.} Due to spoof attacks variations and the possibility of fabricating new attack types, a model may require to be retrained anytime new attack is presented. To overcome this issue, some FAS studies have considered liveness detection as a  zero-shot or few-shot problem. Thus, liveness models can detect unseen attack types or be more stable in detecting previously known attacks.

\textbf{Generative Models.} Typically, FAS methods rely upon learning discriminative features to detect spoof attacks, while generative models try to estimate spoof traces of an input. 
By extracting spoof traces, not only it is possible to detect spoof attacks based on them, but also synthesize new spoof images.

\textbf{Neural Architectural Search.} Solving tasks using deep learning depends on finding an optimal set of parameters in a deep neural structures. Usually, this process takes place through numerous trial and errors and based on previous experience, but NAS automates this procedure to achieve the final structure. 

Moreover, we explain major public datasets that are commonly used in PAD studies in detail and mention their limitations and advantages. The main contribution of this survey can be listed as follow:
\begin{itemize}
\item In-depth review of recent deep-learning-based PAD methods ( focusing on works published since 2017.
\item Providing a taxonomical categorization of all recent works.
\item Thorough review of significant public datasets.
\item Suggesting research areas for future works considering the current deficiencies.
\end{itemize}

\subsection{Organization of the Paper}
In section~\ref{sec:approaches}, we introduce each aforementioned category and explain how they resolve existing issues in PAD, and we list all related recent works of each category by the time of their publication to demonstrate how they have evolved through time. In section~\ref{sec:datasets}, we give through details about the structure and creation process of all datasets in chronological order and point out their advantages over prior ones. In section~\ref{sec:results}, we introduce evaluation metrics for FAS systems and present quantitative comparison between reviewed methods based on major datasets used for benchmarking. In section~\ref{sec:limitations}, we consider given results in section~\ref{sec:results} to point out the limitations and possible directions of future research; finally, we give our conclusions in section~\ref{sec:conclusion}.


\section{Approaches}\label{sec:approaches}
A comprehensive categorization of retrospective research in deep FAS not only helps new researchers to achieve an inclusive review of available methods but also help them to better understand task specific challenges and develop new methods corresponding to their needs and requirements. Various factors and benchmarks can be used to achieve this classification. For instance, \cite{yu2021deep} reviews existing deep learning based FAS methods using commercial RGB camera according to hybrid methods (combination of handcrafted and deep-based features), common deep learning based methods, and methods used for increasing model's generalization capacity on unseen domains and unknown attacks. \cite{ming2020survey}, on the other hand, provides a hierarchical taxonomy first according to capturing device (RGB commercial camera and specific hardware) and then relative to feature types and methods.

The assortment presented in this paper resembles the mentioned ones, however we attempt to focus on novelties and the corresponding pros and cons compared to other methods. Accordingly, we use the following factors as criteria to categorize RGB camera-based deep FAS methods:
\begin{enumerate}
    \item Type of auxiliary information used as complementary information 
    \item Intrinsic liveness-related features
    \item Machine learning techniques 
\end{enumerate}
 
In the following, we introduce each of these categories and then discuss methods falling into them. 

\subsection{Auxiliary Information Tasks}\label{sec:aux_inf}
Similar to most of the machine learning problems, FAS networks suffer from overfitting ensuing malperformance on datasets different from those seen during the development phase~\cite{lucena2017transfer, rehman2017deep, yang2014learn, gan20173d, alotaibi2017deep}.
In order to alleviate this problem, ~\cite{desai2019face, li2019face, qu2019shallowcnn, zhang2020face} tried to feed deep features, traditional low-level information, as well as hybrid features to a machine learning classifier such as SVM. Unfortunately, these methods may capture only discriminative features like screen bezel or display frame instead of liveness-related features and faithful spoofing patterns. 

Another well-known solution is to add the supervision of an auxiliary signal to the training procedure~\cite{sun2020face, liu2018learning, kim2019basn, yu2021revisiting, nowara2017ppgsecure}. Depth map, Reflectance, Remote PhotoPlethysmoGraphy (rPPG) signal, and Binary mask, are several well-known auxiliary information in the context of face anti-spoofing research. 

\subsubsection{Depth and Binary Maps}
From the spatial perspective, depth map is one of the features that is clearly discriminative between bona fide and spoof images. Intuitively, some part of face is closer to the camera than the other regions in a genuine sample while the depth map of a 2D spoofed face is flat or planar and consequently all pixels have the same depth to the camera. Such discriminative signal brought by depth maps can provide a proper supervision for FAS network training, and guides the network to capture more informative features. 

To leverage signal, first step is to generate depth maps of genuine faces using a depth map estimator \cite{feng2018prn} as ground truth and to consider zero maps for spoof samples. Supervised by a pixel-wise loss function ~\cite{kim2019basn, liu2018learning, yu2020face}, FAS networks can generate an estimation of facial depth map.

This type of supervision has widely been employed in PAD studies. 
For instance, Yu et al.~\cite{yu2020face} employ depth to learn detailed cues effectively. 
More specifically, they restated FAS as a \emph{material perception} problem in which physical spoofing carriers (e.g., paper, glass screen, and mask) are the main distinguishing factor between live and spoof faces, resulting to a new formulation for PAD.
Accordingly, the authors employed bilateral filters that plays a vital role in representing macro- and micro- cues for various materials. Although bilateral filtered (i.e., bilateral base and residual) images can be fed directly to the network, they might result in information loss with less representative features. Moreover, it would be inefficient to learn multi-level bilateral features when bilateral filter is only applied on the input space.
In order to mitigate this issue, Deep Bilateral Operator~(DBO) is  applied on deep features instead of original image. This operator is used in Bilateral Convolutional Network (BCN) to extract intrinsic patterns via aggregating multi-level bilateral information.
Furthermore, in order to achieve more optimal features for fine-grained material-based FAS task, Multi-level Feature Refinement Module (MFRM) is exploited after BCN, aiming to refine and fuse the coarse low-mid-high level features from BCN via context-aware feature reassembling.
Despite the performance boost achieved by psudo-depth maps supervision, the conventional depth maps have the following drawbacks ~\cite{sun2020face, liu2018learning}:
\begin{enumerate}
    \item The process of synthesising depth maps for training is costly.
    \item Depth maps are not accurate for partial spoof attacks.
    \item It may fail to detect attacks with 3D properties such as high-quality 3D mask attack, makeup attack, Mannequin, etc.
\end{enumerate}
To resolve these issues, binary maps have been proposed as a better substitution for depth maps, as they are easier to generate, and allows researchers to generate accurate binary masks for partial attacks.
For example, George et al.~\cite{george2019deep} used pixel-wise binary supervision provided by binary masks, i.e. zero map for spoof face and one map for genuine one. Thus, the process of producing auxiliary map would be more cost effective. 
Sun et al.~\cite{sun2020face}, on the other hand, utilized more elaborated process to produce binary maps for bona fide samples in which the background regions are masked out and do not have any effect on the final decision. More specifically, for a given frame consisting of a single face, the genuine foregrounds, spoofed foregrounds, and the background scene are considered as positive, negative, and undetermined, respectively.  
The proposed method, Spatial Aggregation of Pixel-level Local Classifiers (SAPLC), is based on the hypothesis that the depth map supervision is unnecessary and local label supervision would be more beneficial due to local nature of face spoofing detection. Moreover, they asserted that face bounding box contains some background pixels that are unrelated to PAD and must be filtered out. To do so, a foreground mask \textit{M} as a binary map is calculated with the existing 2D facial landmarks, representing the genuine foregrounds of the face.

To fully exploit depth maps or binary masks, Yu et al. ~\cite{yu2021revisiting} introduced pyramid supervision to provide richer multi-scale spatial context for fine-grained learning, where the original pixel-wise label is decomposed into multiple-spatial scales for supervising multi-scale features. Therefore, the model learns both global semantics and local details from multi-scale features, resulting in a more robust system. To verify this technique, they conduct extensive experiments consisting of ResNet50 or Central Difference Convolutional Networks~(CDCN), binary masks or facial depth maps, and different configuration for number of intermediate supervisions and their corresponding scale size.

\subsubsection{Remote PhotoPlethysmoGraphy (rPPG) signal} \label{sec:rppg}
Pseudo-depth map and binary mask signals are both extracted based on the spatial information and ignore the temporal features providing crucial information specifically to tackle static attacks. One of the prevalent auxiliary information that take into account the temporal information of videos is Remote PhotoPlethysmoGraphy (rPPG) signal. In a live video, heart pulse can be detected from the micro color variation of skin due to the blood flow. The absence of this information in static attacks (e.g. photo attacks) and also its degradation in video replay attacks make the rPPG an informative signal that can be exploited independently for PAD ~\cite{nowara2017ppgsecure, liu20163d} or as a complementary signal to cover the deficiencies of other information sources~\cite{liu2018learning}.

rPPG can be exploited as an adequate explanatory information for FAS. For instance, ~\cite{nowara2017ppgsecure} and ~\cite{hernandez2018time} feed only rPPG signals of the input videos to an SVM and Random Decision Forest to classify them as real or spoof. 
Another approach is to use rPPG signal in deep networks as an auxiliary supervision information similar to depth and binary maps. For instance, Liu et al. ~\cite{liu2018learning} use combination of depth map and rPPG as auxiliary supervisions. The proposed network comprises a CNN-based module supervised by depth map to extract spatial information and an RNN module supervised by rPPG to process temporal features. Moreover, to face the problems associated with poses, illuminations, and expressions, known as (PIE), the authors assume that the rPPG signal extracted from videos with no PIE variation can be used as the `ground truth' label for other videos belonging to the same subject since other videos were recorded in a short span of time (less than 5 minutes). Moreover, for spoof videos the rPPG signal is set to zero.
The proposed network benefits from both supervisory signals through a non-rigid registration layer that is embedded in the network to handle the motion and pose changes for rPPG estimation. By utilizing the estimated depth map, this non-rigid registration layer aligns the feature maps learned by CNN module and helps the RNN module to track and learn the changes in the same facial area across time. 

Although lots of research was motivated to use rPPG signal, it suffers from several important drawbacks~\cite{liu2018learning}:
\begin{enumerate}
    \item Variations in pose and face expression have negative impact on the quality of rPPG signal, making it hard to track the specific face area for measuring intensity changes.
    \item Illumination is another factor that overshadows the quality of rPPG signal, since changes in the ambient lighting affect the amount of light reflected from the skin.  
    \item High quality replay attacks could generate rPPG signals similar to live samples.
\end{enumerate}

\subsubsection{Face Reflectance}
A valuable auxiliary information is reflenctance which is informative especially for replay attacks where the surface of spoofing medium causes reflection artifacts in recaptured images. While the surface of genuine faces rarely contain reflection artifacts, Reflectance, often caused by the surface of spoofing medium in recaptured/replayed videos, is an attractive auxiliary information for FAS, especially in replay attacks.

Reflectance map is usually used along with other auxiliary signals such as depth, as a part of a bipartite auxiliary supervision, to enrich feature representation and improve the generalization capabilities of models~\cite{pinto2020leveraging, yu2020face, kim2019basn}(We have already discussed these works in details previous sections, so that we avoid repetition.). 

According to the reflection caused by Presentation Attack Instruments(PAIs), some researchers attempt to intensify the reflection effect on subjects face, resulting in more vivid difference between bona fide and live faces. This device-based auxiliary information is based on the reflection of light generated and cast by a light source. The difference between live and genuine faces response to the emitted light is a promising cue to detect spoof attacks. Reviewing FAS methods deploying this technique ensues better understanding.

Following this, Tang et al.~\cite{tang2018face} introduced a new liveness detection protocol called \textit{Face Flashing}. In this technique, the screen of the capturing device flashes a randomly generated light reflecting back into the subjects face. During this process, the subject's video containing the immediate reflected light from the face is collected through the activated camera. This process is repeated several times to increase the robustness and security of the system. Moreover, the location of the lighting area plays an important roll in spoof and live videos differentiation; more specifically, the reflected light captured by the camera in a certain pixels is proportional to the incoming light of the same pixel. Using this phenomenon, in the genuine videos, the emission of the incoming light and the light reflection occurs at the same speed, so that the changes can be regarded as simultaneously. On the other hand, the adversary’s screen can hardly be synchronized with the capturing device's screen. In other words, the adversary cannot interrupt the screen and let it show the latest generated response before the start of next refreshing period. Consequently, the main difference between live and spoof is the delay between the exposed light and the captured frame.  

Liu et al.~\cite{liu2019aurora} likewise proposed a method based on light reflection, named Aurora Guard (AG). This method utilizes light reflection to impose two auxiliary information: light parameter sequence and depth map. In this method, the capturing device acts as an extra light source and casts dynamic light specified by a random light parameter sequence (i.e. light CAPTCHA  that specifies light hues and intensities). Based on the recorded video, normal cues are extracted from facial reflection frames by estimating reflection increments on subjects' face. Detailedly, the light is reflected and scattered from face as a diffuse reflectors and then is perceived as the final imaging in the camera.  Given images containing reflection on the object surface, the magnitude variations among different images is measured, under the assumption of Lambertian reflection~\cite{tan2010face} model and weak perspective camera projection. 
These normal cues are also used as the inputs of an encoder-decoder network to generate depth map that is fed to a classifier discriminating between 3D genuine face and 2D presentation attacks. Further, to increase the security and robustness of the system, the light CAPTCHA is also decoded back from normal cues using a regression branch.

\subsection{Intrinsic Features}\label{sec:intrinsic_features}
Intrinsic characteristics and differences between live and spoof images are inevitable factors that researchers take into account when proposing new methods. These features and clues can lie in different color spaces or various maps extracted from the raw image. Various methods and networks attempt to utilize these peculiar features to differentiate live and spoof samples. 
Accordingly, FAS networks can be lumped into two categories. The first group focuses on spatial or textural information of a single frame while the second one utilizes the temporal information in the sequence of frames along with textural clues in each frame. In the remainder, we consider the first group as spatial and the second group as spatio-temporal features and accordingly explain different FAS networks in each group.

\subsubsection{Spatial Features}
In face texture analysis, conventional feature extractors and CNNs analyze the image from different perspective and offer various advantages. For instance, deep models present robust features extracted from data whilst handcrafted ones have great ability in characterizing the image.
Li et al.~\cite{li2019face} proposed a hybrid method to benefit from both features. First, they fine-tuned a pre-trained face recognition model named VGG-face with anti-spoofing datasets. Then, they derived LBP features for each convolutional feature map and its corresponding histogram. Finally, the concatenation of feature histograms is given to an SVM classifier. Further, to investigate the effect of various color spaces on the final performance, the authors also processed the input image in RGB and HSV spaces as well as YCbCr. 

Conventionally, face image in RGB space is usually fed to the FAS networks but Pinto et al.~\cite{pinto2020leveraging} tried to use reflectance, albedo, and depth maps as inputs to extract more meaningful information. More specifically, they used a technique called Shape-from-Shading (SfS), which aims to estimate the shape of an object based on the shade information presented in the surface. SfS can estimate practical information like depth, reflectance, and albedo from a single RGB image captured without using any additional sensors. 
Technically speaking, SfS maps are calculated for each channel of the RGB input image at once and then concatenated to form a nine-channel input tensor. Assuming that the reconstructed surface from a spoof face contains synthetic patterns, the CNN can benefit from them to detect spoof attacks. Reflectance, depth, and albedo can likewise be fed separately as the input to network and each can be considered as an independent classifier or the final decision can be based on major voting; however, the results confirm the superiority of multi-channel input tensor.

Instead of concatenation of different maps, it is also possible to feed each of them to the same network and then fuse the corresponding features to achieve more informative ones for final classification. Cheng et al.~\cite{chen2019attention} deploy this method and presented a Two-Stream CNN (TSCNN) consisting of two color spaces: RGB and Multi-Scale Retinex (MSR) spaces. The RGB stream provides the model with detailed facial information, while MSR which is invariant to illumination, can capture high-frequency information effective for spoofing detection. In this network, ResNet18 is used to extract features from both feature spaces and then on top of that, attention-based fusion is adopted to weigh features adaptively.

An alternative approach of processing various spaces and maps with the same back-end network is to feed the same input to various feature extractors to extract different types of clues from the same input.
For instance, Atoum et al.~\cite{atoum2017face} proposed a two-stream network, wherein both streams ingest face image in HSV, YCbCr, or their concatenation formats. In this network, a patch-based stream extracts discriminative features from randomly selected patches of original face image.
An influential characteristic of patch-based processing is that it preserves the original face information. In other words, since the feature extractor does not process the whole face image at once, it does require to resize the original one which cause losing the discriminative information. Moreover, this patch-based method can detect the spoof pattern independent of face region and extracting random patches from the original image can results in more training data. This patch-based CNN produce spoof scores ranging from 0 to 1 for every patch. 
On the other hand, the other stream first learns the face's holistic depth map and then extracts features from this map using another CNN module. These features are then fed to an SVM for final classification.
Finally, the patch scores and the SVM classifier results are fused for final decision.

Despite the good performance of two-stream networks for feature extraction, their training procedure is both challenging and time consuming. Transfer learning can be an appropriate solution to overcome this challenge. For instance, Muhammad and Hadid~\cite{muhammad2019face} have proposed a network in which the face image is first fed to ResNet-50 and Resnet-101 both trained on ImageNet, and in the second step, to achieve more informative features,  Principal Component Analysis (PCA) and Canonical Correlation Analysis (CCA) are applied on the output features for dimensionality reduction and feature fusion, respectively. Finally, the output of the second step is fed to an SVM as the final classifier trained using one-versus-all strategy.

\subsubsection{Spatio-temporal Features}

The methods described so far mainly focus on textural information represented in static images and ignore crucial motion related information that can help to boost the FAS accuracy, especially for static attacks~(e.g. photo, photo replay, print attacks, etc.) lacking any dynamic information in the subject's face. More specifically, eye blinking and movements of lips and other facial components result in various spontaneous local facial muscle movements that are too subtle to be easily detected either by human eyes or by existing face anti-spoofing methods based on hand-crafted dynamic textures as well as spatial deep features~\cite{shao2018joint}. In order to address this challenge, exploring the temporal information of genuine faces is a good solution to include motion in decision making process.

A simple approach is to stack several frames and feed them as a single input to a CNN extracting temporal local features.  Tang et al.~\cite{tang2018fusing} deploy this strategy along with exploring various color spaces and patch based analysis. In particular, gray-scale frames are stacked together to form an \emph{n}-channel input and consequently the extracted features include both spatial cues and local temporal motion between consecutive frames. The authors also used CNNs to learn high-level color-based features of a frame in RGB, HSV, and YCbCr color spaces, since HSV and YCbCr color spaces offer more discrimination for face anti-spoofing as compared to the RGB counterpart. In order to improve the performance, randomly selected patches are processed to include local spacial information in the final decision. The results show that the combination of these three types of features improves the overall EER, although it would be beneficial to evaluate features combination, such as temporal and patch-based features, in order to evaluate influence of each module.

In another approach, according to the motion divergences between real and fake faces, Liu et. al~\cite{liu2019static} designed a multi-modal method to increase the exchanges and interactions among different modalities, aiming to capture correlated and complementary features as well as temporal information.
They introduce a Partially Shared Branch Multi-modal Network (PSMM-Net) wherein there is a branch for three modalities RGB, Depth, and Infrared (IR). Each branch of PSMM-Net is composed of a Static and Dynamic-based Network, called SD-Net, ingesting the RGB (Depth or IR) image and its corresponding dynamic image calculated by rank pooling technique. Rank pooling captures the temporal ordering of a video and learns a linear ranking machine based on the contents of the frames. For SD-Net, rank pooling is directly applied on the pixels of RGB (Depth or IR) frames so that the final dynamic image is of the same size as the input frames.
The backbone feature extractor is based on ResNet-18 in which features of three mention modalities are fused at different stages.

Although frames stacking and rank pooling are two techniques widely used in context of video processing, another common approach is to capture temporal information by Recurrent Neural Networks, such as Long Short-Term Memory(LSTM). LSTM along with CNNs can combine both spatial and temporal features resulting in better class discrimination. For instance, Tu et al.~\cite{tu2017ultra} extracts textural features using ResNet-50 pre-trained on ImageNet and then feed them to an LSTM to learn long-range temporal dependencies from sequence of frames.
Yang et al.~\cite{yang2019face} likewise utilized LSTM to capture underlying temporal information. Specifically,  they propose a Spatio-temporal model named Spatio-Temporal Anti-spoofing Network (STASN), which consists of three parts: Temporal Anti-spoofing Module (TASM), Region Attention Module (RAM), and Spatial Anti-spoofing Module (SASM). In TASM, features extracted by ResNet-50 are given to an LSTM. Moreover, since not all regions of an image contain discriminating information for FAS, RAM extracts important local regions and delivers them to SASM to decide upon local region liveness.

As we stated before, long-range spatio-temporal context can help to boost FAS systems performance; however, classical LSTM and GRU~(Gated Recurrent Unit) ignore the spatial information in hidden units~\cite{chung2014empirical}. In order to alleviate this problem, Wang et al.~\cite{wang2020deep} proposed ConvGRU that propagates the long-range spatio-temporal information and considers the spatial neighbor relationship in hidden layers. 
The proposed network dedicates an independent stream to process current frame $\emph{t}$, one for future frame, i.e. \emph{t + $\Delta$t}, and one stream for communication of first and second ones.
The first and second streams rely on Residual Spatial Gradient Block (RSGB) for feature extraction.  RSGB is based on the difference between gradient magnitude responses of the living and spoofing faces to capture their discriminative clues. More specifically, using shortcut connections, RSGB aggregates the learnable convolutional features with gradient magnitude information in order to extract rich spatial context useful for depth regression task.
The output of these streams is an estimation of face's depth map.
The third stream is based on Short-term Spatio-Temporal Block (STSTB) that extracts generalized short-term spatio-temporal information and fuses different types of features together. The STSTB features are then fed to the ConvGRU block that likewise outputs a depth map used to refine the current depth map.

Temporal convolution, an alternate method for sequence processing, has been applied to video analysis and demonstrated promising results~\cite{nan2021comparison}. Huang et al.~\cite{huang2020deep} deployed a sub Temporal ConvNet to process liveness related features temporally.
Most of the anti-spoofing methods have a hard time detecting spoof attacks when fake cues lies in the low level image pixels. Consequently, the proposed method employs multi-frame frequent spectrum images to pass low-level information through the network to the classifier. 
The main idea stems from the comparison of frequency responses of fake and live images, calculated by Fourier transform. More specifically, there will be color distortion, display artifacts or presentation noise in fake images that are expressed in high frequency domain of their spectral images. 
Thus, a \emph{frequent domain spectral} image is fed to the network, as an additional input, to increase the representational capacity of spoofing features.
The proposed network includes two streams; a ConvNet stream that individually processes the 6-channel stack of RGB and HSV counterparts of each input frames and estimates the corresponding depth map. Following that,  there is a sub Temporal ConvNet that ingests the sequence of depth maps and generates the feature representation in both spatial and temporal domains. In the second stream, a shallow network generates frequent temporal feature representation in low semantic level following by a Temporal FreqNet to evaluates the temporal differences across the frame sequence automatically.
Furthermore, by sampling blocks in spectral images and extracting spoof patterns, the same corresponding area in the live face can be replaced to synthesise more fake samples.

\subsection{Transfer Learning}\label{sec:machine_learning}
In transfer learning, the goal is to exploit the learned knowledge from a previous task in our new but related problem, such as using the weights of a CNN trained for car classification to detect trucks. However, in FAS, we are more interested in subcategories of transfer learning that transfer knowledge of one task from one domain to another. Specifically, we review domain adaptation and knowledge distillation, which the former exploits source domains for target domain to acquire generalized feature space, and latter leverage teacher-student framework achieve the same goal. 

\subsubsection{Domain Adaptation}\label{sec:domain-adaptation}

When assessing the performance of an FAS system, generalizability of the learned feature space is an essential quality. The trained model might achieve exceptional accuracy in intra-dataset evaluations but fail on unseen distribution. This lack of generalization raises various security concerns since not all the attacks done by the imposter or all types of capturing devices can be included in datasets used in development phase.

As a solution adopted from transfer learning method, Domain Adaptation~(DA)~\cite{farahani2021brief} tries to reduce disparity between source and target domain, improving model's performance on unseen data.  
This technique is applicable when the training data does not accurately have the same distribution as test data which is quite often in real-world applications and results in performance degradation.

In face anti-spoofing methods, according to the availability of target labels, DA can be used in two configurations to find a shared feature space between seen and unseen data distribution: unsupervised ~\cite{li2018unsupervised, zhou2019face, mohammadi2020domain, wang2019improving, wang2020unsupervised, jia2021unified, qin2020one, quan2021progressive,panwar2021unsupervised}, semi-supervised~\cite{tu2019deep, jia2021unified}.
These two are of particular importance because of the availability of partially labeled target data~(semi-supervised) or lack of full annotation in target domain~(unsupervised) for real-world applications.

In order for DA to reduce the distribution distance between source and target domains, several distance measures can be deployed, such as Maximum Mean Discrepancy~(MMD)~\cite{li2018unsupervised, tu2019deep, zhou2019face, li2020face} and KL divergence~\cite{mohammadi2020domain}. MMD is usually used to find a shared feature space between source and target domains. For instance, Li et al.~\cite{li2018unsupervised} leveraged MMD for DA in FAS. In particular, they presumed that the features extracted from one particular domain form a compact distribution, and consequently unsupervised DA will leads to an improvement over the generalization ability of face spoofing detection by minimizing the MMD. The second assumption is that outliers result into a sensitive classifier on source domain and a significant distance between target and source distributions. Thus, in order to achieve better results through DA, the authors removed outliers by first applying cross-validation to acquire an accuracy value (acc) using libSVM \cite{chang2011libsvm} and then removed (1-acc) percent of the total samples with the smallest weight values.
Their main goal in this paper is to learn a mapping from source domain to the target one by minimizing the MMD. This is achieved by employing subspace alignment that learns a transformation from the source subspace to the target subspace. Using this transformation, the source domain
is mapped to target domain and finally the classifier is trained based on the transformed source domain data.

MMD is usually applied on the last fully connected layer of the network~\cite{li2018unsupervised, tu2019deep, li2020face} which according to \cite{qiu2018deep}, increase the distribution distance between source and target domains. Additionally, applying MMD on the last fully connected layer does not consider the impact of other layers on the data distribution. In order to address this issue, Zhou et al. \cite{zhou2019face} suggested a multi-layer MMD~(ML-MMD) for unsupervised DA to combine representation layers with classification layers. ML-MMD can bridge the distribution distance by considering both marginal and conditional distributions of multiple layers. In the proposed network, there are also two skip connections after the first two pooling layers, resulting in more detailed features.
Classification layer is trained on the source data, while domain adaptation layers use data from both domains. The network's final loss function is a combination of both classification loss and MMD distance.

Mohammadi et al.~\cite{mohammadi2020domain} likewise proposed a method that utilizes features at different depths to better generalize on the target data. They hypothesized that in a CNN trained for PAD task, the initial layers extract features that generalize better to the target dataset while deeper layers focus on domain specific characteristics. 
Moreover, in their proposed method, they use Kullback–Leibler Divergence~(KLD) to measure domain shift at a given layer in the feature extractor. The method helps to retain filters that are activated for both source and target domains. As such, this technique is similar to feature selection wherein \emph{N} percent of filters in layer \emph{L} are pruned and then the next layer \emph{L+1} is re-trained on the source data using the same classifier. This approach will force the final classifier to make prediction according to the robust features shared between source and target distributions.

In aforementioned methods, we have reviewed domain adaptation methods that have employed a specific metric~(MMD,KL divergence) to circumvent domain shift and achieve a generalizable model. Nevertheless, Wang et al.~\cites{wang2019improving} introduced an Adversarial Domain Adaptation~(ADA) approach to build the target domain PAD model. The training procedure of the proposed networks consists of two main stages. First, a source model is pretrained on the source dataset and is optimized with triplet loss, to handle within-class diversity that particularly happens in spoof category, where, k-Nearest Neighbors (k-NN) is used as the classifier due to the fact that live and spoof features form two different clusters under a shared embedding space.
In the second stage of the development procedure, the adversarial adaptation is used to learn a target model such that under the embedding space the discriminator cannot reliably distinguish between source and target encoders’ features.

Later, the authors also applied two improvements to their proposed method in order to achieve better performance on target data~\cite{wang2020unsupervised}. First, they added center loss to the existing triplet loss to extract more discriminative features. Second, a new module named Disentangled Representation network (DR-Net) is employed to disentangle the features in the shared feature space of source and target domains, so that features become more domain-independent. To do so, they reconstructed input images from the shared feature space using source and target decoders. This process is supervised via L1 loss and feature-level loss. 
The convolutional layers in encoders are used to minimize the distance in feature level, and due to the success of perceptual loss \cite{wang2020unsupervised} in style transfer and super-resolution, this feature-level loss is added to the final loss as reconstruction loss .

Although the assumption of unsupervised DA is reasonable, especially in FAS, Tu et al.~\cite{tu2019deep} developed an approach based on the fact that in real-world scenarios, it would be possible to acquire a limited amount of labeled data from target domain. They employed a `Two-Half' strategy in which one half of each training batch belongs to the source domain and the other half to the sparsely labeled target domain.
In training phase, AlexNet~\cite{krizhevsky2012imagenet} is trained on the source domain data to learn discriminative deep features, through a cross-entropy loss. Moreover, a kernel based metric is used in order to measure  distribution distance between source and target domains.They demonstrated that applying a suitable kernel on the input data results into more convenient feature space and easier quantification of the distance between various distributions. Tu et al. selected a mixture of Radial Basis Function~(RBF) kernels to compute MMD between source-domain and target-domain samples.
The MMD can be more specialized according to the material of their spoofing surfaces (e.g. print paper, video screen, mask).
Using this intra-modality approach, the MMD calculates the distance between various modalities in source and target domains.
The final loss function comprises the classification loss on the labeled source-domain data and the domain loss between the source data and target data.

Alternatively, Jia et al.~\cite{jia2021unified} proposed a DA method that can be applied in both semi-supervised and unsupervised manner according to target domain. Their structure consists of two main components; Marginal Distribution Alignment~(MDA) module and Conditional Distribution Alignment~(CDA) module. The former, i.e, MDA, trains the source and domain generators in an adversarial fashion against a domain discriminator to find a domain-independent feature space. The loss function for this procedure is the conventional minmax loss~\cite{goodfellow2014generative}, but instead of training the generator and discriminator iteratively~\cite{wang2020unsupervised, wang2019improving}, they add a gradient reversal layer~(GRL) layer~\cite{ganin2015unsupervised} before the domain discriminator to train them simultaneously. The latter, i.e, CDA, simply minimizes the distance between class centroids of both domains; thus, embeddings of each class from both domains become closer together.
Moreover, to further refine the classifier, entropy minimization principle~\cite{grandvalet2005semi} has been applied on unlabeled target data to encourage low-density separation between classes of the target data. Additionally, normalization on the outputs of feature generator and classifier is used to have a normalized and cosine-similarity-based feature space to improve the generalization. In semi-supervised mode, the loss is aggregated via cross-entropy loss over labeled data~(source data, and some target data), adversarial loss, entropy loss, and semantic alignment loss~(CDA loss). When no labeled target data is available, the semantic alignment loss is eliminated and the target data is used in unsupervised manner.

In most recent approaches in face-anti spoofing, meta-learning plays a major role since it can be used to mimic real-world scenarios in which target data is not available during training, aiding the model to become more generalizable. In this direction, Qin et al.~\cite{qin2020one} carry out domain adaptation by introducing One-Class Adaptation Face Anti-Spoofing~(OCA-FAS) in which they train a meta-learner for OCA tasks to learn adaptation using only genuine faces; the used deep networks include a feature extractor, binary classifier, and a regressor. They also present a novel Meta Loss Function Search~(MLS) strategy to aid the meta-learner to find loss structure for OCA tasks. The training process includes two stages: one-class adaptation and meta-optimizing stages. 
In the first step, the meta-learner adapts its weights with only some live faces of the task via a supervised  OCA loss, which is the weighted summation of following three loss functions: binary-cross-entropy loss, pixel-wise loss, and simplified deep SVDD loss~\cite{ruff2018deep}, which calculates variance of extracted features. In the next stage, the modified meta-learner is optimized using spoof faces and remaining live faces in the OCA task to learn the weights of feature extractor, binary classifier, regressor, coefficients for each term of the OCA loss, and the learning rate in the first stage. Since in the meta-optimization stage both live and spoof data is available, the deep SVDD loss function is eliminated. During test, the output of the model is the average of binary classifier and the regressor.

Thus far, we showed how target domain can be used in unsupervised or semi-supervised manner in training process. However, Wang et al. \cite{wang2021self} propose a meta-learning based adaptor learning algorithm in which model is only trained using source domain data to initialize an adaptor that can adapt itself to test domain at inference. The propose network comprises of five modules: a feature extractor (\textit{F}), a classifier (\textit{C}), a depth estimator (\textit{D}), an adaptor (\textit{A}), and an autoencoder (\textit{R}). The adaptor is a $1\times1$ convolution and is connected to the first convolution layer in C in residual fashion. The training process has two main stages: meta-train and meta-test stages. 

In the first stage, from \textit{N} available source domains, one is randomly selected as meta-train domain and another one is selected as meta-test domain. Then, \textit{F}, \textit{C}, \textit{D}, and \textit{R} are trained using meta-train domain, the input of the \textit{R} is the features maps of classifier after first activation layer. 
Afterward, the adaptor is added to the network to adapt the features of \textit{C} using target domain (meta-test domain). Since target domain labels are not available in real-world applications, they train the adaptor in unsupervised manner using following three loss functions: reconstruction loss, entropy, and orthogonality. The latest one is used to prevent feature mode collapse and they impose orthogonality on adaptor using Spectral Restricted Isometry Property regularization \cite{bansal2018can}. During training the adaptor, except the adaptor itself, the rest of the model parameters are fixed. In meta-test stage, they only calculate the adapted classifier and depth loss on meta-test domain. Once the model is trained, first, the adaptor is optimized using new unseen data in unsupervised manner, then all model parameters are fixed and inference takes place on target data.

Commonly, each dataset is considered as a separate domain in domain adaptation or generalization frameworks. Nonetheless, we may not have access to ground truth domain labels of data in real-world scenarios. Thus, Similar to~\cite{chen2021generalizable}, Panwar et al.~\cite{panwar2021unsupervised} argue that domain adaptation approaches should not depend on domain labels. Their work separates the dataset only into source and target domains, and the target domain contains unseen spoof attacks. The training process takes place in four stages:
\begin{enumerate}
\item Training source network using ground truth labels (live/spoof) by cross-entropy loss. 
\item Initializing target network with the pre-trained source model weights, and training it for domain alignment in adversarial manner using domain discriminator. During training, they use a memory-based source-to-target knowledge transfer technique, which uses the means of live and spoof features from the source domain to enhance the learned representations in the target network, and the source network parameters are fixed during the training of the target network.  
\item Training Domain Specifier network (DSN) to learn domain-specific attributes from both source and target domain. Trained DSN is later used to rank samples' difficulty from both domains using L2 distance.
\item Training target network based on DSN ranking to further improve the representation for FAS.
\end{enumerate}

All the mentioned works are formed based on the hypothesis that an adequate amount of labeled data is available for training, but labeling data is a laborious and time-consuming process. Therefore, a framework capable of learning through a small amount of labeled data would be very convenient. Quan et al. \cite{quan2021progressive} 
propose a progressive transfer learning method for FAS in which training begins with a small amount of labeled data and model iteratively assigns pseudo-labels to the remaining unlabeled data. During label assigning, only samples with high confidence are selected, and they employ temporal constraint mechanism to exploit other frames of a same video to calculate model confidence for a specific frame. Specifically, for each unlabeled frame, the mean confidence of all frames of its video and the confidence of that frame are averaged. Additionally, if the number of labeled spoof and live data is not larger than an adaptive threshold in an iteration, the algorithm will turn into the next iteration to gather more data without updating the model to further guarantee model stability.

Following steps summarize the training procedure for inter-dataset case, which only utilizes source domain data:
\begin{enumerate}
\item training the model with a limited number of labeled data from the source domain.
\item assigning pseudo-label to unlabeled data based on the model’s predictions for samples with high confidence.
\item training the model with manually annotated labels and samples with pseudo-labels.
\item repeating 2 and 3 steps until all data is labeled.
\end{enumerate}
For inter-dataset scenario, first, the previously trained model on source domain is used as initial weights. Then, same as intra-dataset scenario, pseudo-labels are assigned to unlabeled data in target domain in each iteration, and the model is trained with source domain data and pseudo-labeled target data. In addition, they propose adaptive transfer to stabilize inter-dataset performance by gradually increasing the contribution of target data and decreasing the contribution of source data in the loss function.

\subsubsection{Knowledge Distillation}
A major challenge of FAS is undoubtedly the lack of enough data for specific scenarios, such as uncommon attacks types, unique acquisition devices, etc. Although multiple datasets are publicly available, to enhance model performance in general, when a model is used in real-world applications depending on the extent of domain shift, the model can be unreliable. 
Additionally, the limited amount of data gathered from your application is not fully exploited when it is used alongside other larger datasets to train a model, and employing the limited data separately can not yield acceptable results.
Therefore, to solve this issue, knowledge distillation can be used in which a teacher network is a complex model and trained on the richer dataset. Then, the knowledge is transferred to the student network that usually has a simpler structure with limited data.

Since acquiring a large amount of data for a new use case is impractical and, in some cases is impossible, li et al.~\cite{li2020face} exploits knowledge distillation to overcome this problem. First, they adopt AlexNet\cite{krizhevsky2012imagenet}, the winning CNN architecture of the 2012 ImageNet ILSVRC competition, for both teacher and student networks. Then, the teacher network is trained with rich-data domain (available datasets for PAD), and they pair rich-data domain with target domain of their application. Rich data goes through the teacher network and the other image in the pair feedforwards through the student network. To distill information from teacher network to student network, they use Maximum Mean Discrepancy (MMD) over the final embeddings of both networks to minimize feature distribution distance. Further, cosine similarity is minimized if labels in the pair match and maximized vise versa, and for better adaptation to the application-specific domain, they apply the cross-entropy loss to the student network.

\subsection{Domain Generalization}
Domain generalization seeks the same goal as domain adaptation, but it has no access to target domain during training. Therefore, domain generalization exploits all available dastasets in source domain to have an acceptable performance against unseen data as well.

Similar to DA, first works that made effort to acquire generalizable features among different domains, employed Maximum Mean Discrepancy (MMD). For instance, Li et al.~\cite{liu2018learning} utilize 3D CNN to extract both spatial and temporal data; the model is trained based on print and replay attacks with two data augmentation techniques. In the first type of data augmentation, spatial augmentation, the face bounding box is shifted in four directions (up, down, left, and right), so part of the face does not appear in the final face image. Moreover, in gamma correction-based augmentation, a gamma correction is applied for each frame of a given video. To provide generalizability, they minimize the feature distribution dissimilarity across domains by MMD, which is added to the cross-entropy loss function as a regularization term. It is worth nothing that they define different domains by camera models and not the datasets.

Most recent works in domain genralization combine domain disriminators and advarsarial learning to force the feature extractor towards a shared space among various domains. Shao et al.~\cite{shao2019multi} train N feature extractors for N source domains that each one is biased to a single source domain; all discriminators are trained by binary-cross-entropy loss with their corresponding extractors. Subsequently, to achieve a generalizable feature space, a multi-adversarial deep domain generalization method is exploited. The primary feature generator of the proposed model competes will all the N discriminators in an adversarial fashion. Thus, the shared feature space among all domains is automatically detected by successfully fooling all the discriminators. To enhance the generalizability, auxiliary supervision of face depth is further incorporated in the learning process.

Additionally, a dual-force triplet-mining constraint is applied to help with the feature space discriminability, which brings two advantages: 1) the distance of each subject to its intra-domain positive (from the same class) is smaller than its intra-domain negative (from different classes); 2) simultaneously the distance of each subject to its cross-domain positive is smaller than to its cross-domain negative.

One drawback of the preceding approach is the requirement of training N feature extractors and discriminators prior to the primary training phase, and as the number of domains increases this would be more resource demanding and time-consuming. To circumvent this obstacle, Saha et al.~\cite{saha2020domain} present class-conditional domain discriminator module coupled with gradient reversal layer. The discriminator classifies the domain of features, conditioned on the live or spoof class. The gradient reversal multiplies a negative constant when backpropagating, which reverses the subjective of subsequent layers. Consequently, the generator will make the feature representation of different domains more compact in feature space. Moreover, there is a binary classification branch to keep spoof and live feature distribution separated. When working with videos, they apply LSTM on the generated feature of each frame to also benefit from temporal information. 

Usually two distributions are assumed in feature space to represent live and spoof samples. However, spoof samples from different domains are very diverse, and assuming the same distribution for all spoof domains can lead to a suboptimal model. Accordingly, Jia et al.~\cite{jia2020single} consider one cluster for all live samples from all domains and one cluster for each spoof domain in the feature space. Domain knowledge is transferred to the feature generator in the same manner as [25] by using GRL before the domain discriminator. Additionally, asymmetric triplet loss is applied, which has the following three benefits:
\begin{enumerate}
\item live embeddings from all sources become more compact.
\item Spoof domains from different domains get separated.
\item The live samples get further apart from spoof samples.
\end{enumerate}
Furthermore, feature normalizatiofon has been applied on the generator output features to constraint all features with the same euclidean norm. In addition, weight normalization and bias elimination on the last fully connected layer make the live and spoof features even further away from one another.

While there have been methods that use adversarial learning for domain generalizations, the direct use of conventional GAN models have been rare. However, GAN models can provide additional data in addition to enabling adversarial learning.  
Wang et al.\cite{wang2020cross} exploits GAN models to learn the distribution of each source domains, and combines the features from different domains to acquire domain-independent PAD features. To this end, they design two modules: disentangled representation module (DR-Net) and multi-domain feature learning module (MD-Net). The DR-Net consists of PAD-GAN and ID-GAN, which are responsible to learn spoof vs live distributions, and subjects’ distributions respectively. Both PAD-GAN and ID-GAN are conventional GAN structure, as \cite{ chen2016infogan, denton2017unsupervised} that use generative models to obtain disentangled representation, and the discriminator has two loss functions; GAN loss differentiates real vs generated images, and classification loss determines class of images (spoof vs loss for PAD-GAN and subject’s ID for ID-GAN). Although it is possible to fuse the features from different domains by concatenating PAD Encoders outputs, they use MD-Net for this purpose. In MD-Net, they use ID encoder of each domain to guide their PAD encoder pair to extract only liveness related features for data from other domains. Finally, a classifier is trained based on the concatenated cross-domain features to be used at inference for unseen data. At inference, features from all PAD encoders are concatenated and given to the trained classifier.

Recently, meta-learning has drew the attention of researchers to achieve robust and generalizable models against unseen data. For instance, Shao et al.~\cite{shao2020regularized} tackle DG in a meta-learning framework and make some adjustment to it to make it more suitable for face anti spoofing. The vanilla meta-learning methods have two issues for face anti-spoofing: 1) if only binary labels are used, since the learning direction is arbitrary and biased, it hinders generalization; 2) dividing multiple source domains into two groups to form aggregated meta-train and meta test in each iteration is sub-optimal since single domain shift scenario is simulated in each iteration. To resolve the first issues, they incorporate facial depth, so the meta-learning is applied over a feature space that is regularized with auxiliary supervised domain knowledge of face anti-spoofing task. Consequently, the meta-optimization can guided and model to capture more generalizable features. To address second problem, the source domain is divided into multiple meta-train, and meta-test sets, and meta-learning is conducted between each pair simultaneously; thus, in each iteration a variety of domain shifts scenario is simulated, which better enables meta-learning to exploit domain shift information.

As we mentioned previously in section \ref{sec:domain-adaptation}, depending on ground truth domain labels cam become troublesome, an in some cases impossible (data gathered from web). To surmount this challenge, Kim et al.~\cite{kim2021domain} divide data into \textit{N} domains using mean and variance of mid-level features of feature extractor and depth estimator, and then apply meta-learning to extract generalizable features. According to previous studies~\cite{li2017demystifying} that shows middle-level features reflect style of data, and Inspired by~\cite{matsuura2020domain} that clusters features to find domains for data, they stack the convolutional feature statics of 5th and 9th layers of feature extractor and the last layer of depth estimator; then, they apply PCA to reduce the stacked features dimension to 256 and cluster them into \textit{N} pseudo domains. Afterward, meta-learning performs three following steps repeatedly: meta-train, meta-test, and meta-optimization. In meta-train, for every randomly selected \textit{N-1} domain, classification and depth regression losses are calculated. Next, for meta-test, they use the updated \textit{N-1} meta-parameters to calculate both losses for the remaining data domain. Finally, in meta-optimization, they conflate the meta-train and meta-test loss values to update the models’ parameters.

Chen et al.~\cite{chen2021generalizable} present a similar appraoch to above-mentioned method~\cite{kim2021domain}, but instead of solely selecting feature from specific layers, they select domain-discriminative features to find pseudo domain labels. They use a stack of convolutional feature statistics, which are the mean and the standard deviation of feature maps from different layers, as the primary domain representation. Additionally, they introduce Domain Representation Learning Module (DRLM) that uses the Squeeze-and-Excitation framework to specify task-discriminative and domain-discriminative features. Further, domain enhancement entropy loss is added to DRLM to improve domain discrimination features, and a depth estimator is applied over task-discriminative features specified by DRLM. To prevent outliers from affecting clustering performance, they introduce MMD-based regularization in the penultimate layer to reduce the distance between sample distribution and the prior distribution. Finally, the training process includes two stages: 1) the pseudo label is determined and assigned by clustering with discriminative domain representations, 2) they train the model by a meta-learning approach.

Although most domain generalization methods exploit datasets' source to find a generalizable feature space, they do not make use of additional information that comes with most FAS dataset. However, Kim et al.~\cite{kim2021suppressing} exploit such information by forcing the feature extractor to capture faithful features of FAS and discarding irrelevant features, such as, identity, acquisition device, environment, etc. The training strategy is comprised of two steps: learning discriminative features while discarding spoof-irrelevant factors (SiFs), improving the encoder discriminability.  For Sifs, they only consider spoof-irrelevant factors that are available in datasets as labels; specifically, identities, environments, and sensors.
In the first step, the encoder \textit{E}, spoof classifier \textit{C}, and secondary spoof classifier \textit{S} are trained. \textit{E} and \textit{C} are trained by minimizing classification loss, and \textit{E} is also trained to maximize SiF classification loss in adversarial manner by a GRL layer between the encoder and SiF discrimination heads. During this step, discriminators are not updated and only pass the gradients to encoder.  In the second step, only the weights of discriminators and intermediate layers are updated and rest of the network parameters are fixed. To strengthen encoders’ discriminability, GRL layer is inserted between secondary spoof classifier and intermediate layers, so the intermediate layers suppress spoof cues, coercing encoders to intensify its discriminative features.  

Unlike almost all previous domain generalization approaches that treat data from different domains indiscriminately, Liu et al.~\cite{liu2021dual} reweights the relative importance between samples to improve the generalization. Specifically, they utilize Sample Reweighting Module (SRM) and Feature Reweighting Module (FRM) to achieve this end. SRM reweights the relative importance of samples in each mini-batch by giving more weight to those samples that are difficult to be distinguished by domain discriminator; because aligning samples with large distribution discrepancies at the early stages of training hinders the model’s generalizability. Moreover, FRM employs reverse weights generated by SRM to suppress the domain-relevant information of large domain-biased samples to help SRM. Training takes place in 4 steps:
\begin{enumerate}
\item updating discriminator \textit{D} with discrimination loss since FRM and SRM modules depend on a fairly good discriminator.
\item the SRM outputs a \textit{W} weight for each sample to emphasize on small domain-biased ones. SRM updates itself via combining \textit{W} and discriminator loss to improve its reweighting.
\item the weights \textit{W} are used to modify the loss function of other modules in the framework, such as, depth loss, binary loss, and discriminator loss, and whole the network (except SRM) updates its weights with these modified loss values.
\item FRM updates itself by combing reverse weights \textit{1-W} with discriminator loss every \textit{K} iterations.
\end{enumerate}

\subsection{Zero- and Few-shot Learning}
Since there are numerous spoof types and attackers can generate novel PAs, if not implausible, it is challenging to cover all attack types during training. As a result, training models often make unreliable decisions when facing new attack types. To address this issue, zero-shot learning and few-shot learning are exploited for FAS. The former, i.e, zero-shot learning, aims at learning both generalized and discriminative features from existing PAs to detect unknown attacks. The latter, i.e, few-shot learning, tries to adapt the previously learned model to new PAs using few samples of new attack types. 

The first attempts on zero-shot face anti-spoofing methods~\cite{arashloo2017anomaly, xu2015learning} have three drawbacks:
\begin{enumerate}
\item \textbf{Lacking spoof type variety}: print and replay attacks are only considered.
\item \textbf{No spoof knowledge}: only live faces disregarding information that spoof data can provide are used.
\item \textbf{Limitation of feature selection}: hand-crafted features like LBP, which is less discriminative than deep semantic features are employed.
\end{enumerate}
To resolve these challenges, Liu et al.~\cite{liu2019deep} provide SiW-M dataset covering five different types of 3D mask attacks, three types of makeup attacks, and three partial attacks; these attacks include impersonation and obfuscation spoofing. 
Based on the assumption that there are some shared and distinct features between different spoof types, they suggest a Deep Tree Network (DTN) to learn discriminating features in a hierarchical manner. Hence, similar spoof types are clustered together in early nodes and separated as we reach leaf nodes. Each tree node contains a Convolutional Residual Unit (CRU) and a Tree Routing unit, while each leaf node contains a Supervised Feature Learning (SFL) Module. CRU is a block of convolutional layers with short-cut connections similar to residual block. TRU is responsible for splitting the data into two subgroups based on the largest data variation. Finally, SFL consists of two branches: fully connected layers to output a binary decision, and the other provides a face depth estimation.

On the other hand, some researchers point out that data-driven models can’t generalize well and have unpredictable results facing new data. Moreover, when adapting to new attacks, they need too many samples, which is impractical. To overcome these challenges, Qin et al.~\cite{qin2020learning} propose Adaptive Inner-update Meta Face Anti-spoofing (AIM-FAS). They solve FAS by Fusion Training (FT) a meta-learner on zero- and few-shot FAS tasks, with an Adaptive Inner-Update learning rate strategy. The meta-learner learns generalizable discriminative features on seen data ,called support set, to have an acceptable response against unseen data. Additionaly, the model performance can be improved on a new PA type when a few samples of that is given to the model. They use a depth regression network named FAS-DR, a fully convolutional network with a depth map as output, and apply Contrastive depth loss (CDL) to get more clear facial depth.

Previously we discussed that, transfer learning is a good technique when annotated data is limited. inspired by this technique, George and Marcel ~\cite{george2020effectiveness} deployed transfer learning to use a pre-trained Vision Transformer model for zero-shot face anti-spoofing task. Transformers model all the pairwise interactions between the components in the input sequence, and thereby, they are mostly used for sequence analysis. To use this method for image classification, an image is considered as a sequence of patches. When using a large set of training data, transformers result in better performance compared to other methods in many vision benchmarks with compensation of computationally expensive training procedure. In order to avoid this issue, the authors used the pretrained weights of a transformer trained for image recognition for PAD and achieved better performance in comparison with other CNNs, such as DenseNet and ResNet. 

In Few-shot learning, models need to learn new PAs by seeing few samples of attacks; thus, exploiting continuous learning would be a promising direction. To do so, Perez et al~\cites{perez2020learning} propose Continual Meta-Learning Face-Pad (CA-PAD) method which takes benefit of continual learning while preserving knowledge of previously seen data using meta-learning. In continual learning gradient alignment is employed to prevent interference during training. Specifically, if inner product of gradients belonging to two arbitrary samples are greater than zero, transfer of knowledge occurs, but interference takes place if the gradient is lower than zero. 

However, continual learning deals with non-stationary stream of data; to address this issue, continual learning is combined with meta-learning via Meta Experience Replay (MER)~\cite{riemer2018learning}. MER augments online learning, so model can be optimized over almost stationary distribution of seen data. In essence, the past data that is stored in a buffer (with capacity of \textit{M} samples) is incorporated in order to not forget previous knowledge while training on new data. Moreover, they employ Reservoir strategy~\cite{ vitter1985random} to replace samples in buffer with M/N probability, where \textit{N} denotes the number of training pairs seen so far. The proposed network structure is composed of three parts in sequential order: fully convolutional backbone consisting of ResNet blocks, depth regressor, and a classifier. Finally, the framework is evaluated using GRAD-GPAD~\cite{ costa2019generalized} (the largest aggregated dataset for FAS).

\subsection{Anomaly Detection}\label{sec:anomaly}
In recent years, researchers have attempted to enhance the face anti-spoofing performance against new PA types or domain shift caused by pose, illumination, and expression (PIE) or sensors variation by adopting different strategies. Usually researchers try to resolve this issue by exploiting large datasets encompassing a diverse range of attack types and finding generalized feature space during training. However, some have suggested that anomaly detection techniques better suit face anti-spoofing task since real samples can be considered as normal and spoof attacks as abnormal. Therefore, with proper modeling of normal samples distributions, any known or unknown spoof samples can be detected due to the deviation from normal distribution. 

Using only normal distribution for anomaly detection often results to inferior performance compared to cases that we have both normal and abnormal distributions. Thus, in recent anomaly detection methods in FAS, spoof attacks representing the abnormal class are also exploited to better find the boundary between two distributions.

To this end, Perez et al.\cite{perez2019deep} aggregate two popular losses which are used in face recognition systems in their approach: triplet focal loss, i.e, a modified triplet loss, and a softmax function named metric-softmax. Considering D\textsubscript{a,p} as a distance of anchor from a positive sample and D\textsubscript{a,n} as a distance of anchor from negative samples, the triplet focal loss divides each distance by a factor and scales it with an exponential function which penalizes hard examples more in comparison to easy ones. The total loss is an aggregation of metric-softmax with triplet focal loss, which the latter acts as a normalizer. During training from a large set of samples, they calculate all possible positive pairs D\textsubscript{a,p} and all possible negative for each positive pair, then negative samples that satisfy $D\textsubscript{a,p} - D\textsubscript{a,n} < m$ are randomly selected. Finally, they suggest two strategies for classifying a new sample:
\begin{enumerate}
\item Training an SVM classifier based on embeddings.
\item Computing the probability of being live given references of live and spoof samples.
\end{enumerate}

Due to variety in PA types, considering a single cluster for all spoof attacks will result in sub-optimal solution for anomaly detection. To address this problem, George et al.~\cite{george2020learning} propose a method in which they compact genuine samples in the feature space and utilize discriminative information from spoof samples without forcing them into a cluster. They use a pretrained LightCNN~\cite{wu2018light} face recognition model and extend it to accept multiple inputs; however, it is possible to only use RGB image as input. Further, they add two fully connected layers on top of the network. They consider the penultimate layer of this network as the embedding and apply the proposed One-Class Contrastive Loss (OCCL) over it.
The OCCL is the modified version of contrastive loss~\cite{hadsell2006dimensionality} which uses center-loss~\cite{wen2016discriminative} instead of pair distance of input samples. For the center loss distance, they consider the distance of samples in each batch from the center of genuine samples. Therefore, compactness constraint is applied only over real class while forcing the attacks away from real ones. The final loss function is the combination of OCCL and binary-cross-entropy loss, which guides the feature space to be discriminative as well. After training the model with the joint loss, one class GMM is used to model the distribution of the compact real class. Finally, at inference, the corresponding feature vector of an input is given to the one class GMM to obtain log-likelihood score of being normal.

Anomaly detection methods usually require a separate classifier to predict the label. However,
Li et al.~\cite{li2020unseen} propose an end-to-end deep framework that only compares the distance of output feature vector to an origin at test time. First, for each input image, they normalize RGB and HSV format and concatenate them together. Then, they feed the concatenated face image to a ResNet18 to transform it to 128 dimensions. Finally, they apply hypersphere loss function on feature vectors which forces normal (genuine) samples towards origin and pushes away abnormal (spoof) ones outside a predefined distance from the origin. They take into account the number of data in each class to adjust hypersphere loss to resolve class imbalance effect. At inference, after extracting the feature vector for an input, they compare the distance from origin with a predefined threshold to detect spoof attacks.

Despite retrospective methods that utilized data from the abnormal class to better found the boundary between live and spoof samples, Baweja et al.~\cite{baweja2020anomaly} endeavor to solve the same issue by only using normal data.

Since considerable differences exists between normal and abnormal classes in some OCC problems, it is possible to solve them by modeling the normal class via zero-centered Gaussian distribution. However, subtle variations among genuine and spoof attacks demands more elaborate strategy, so an adaptive mean estimation is suggested to generate pseudo-negative data during training. Their methods is comprised of two main parts: a feature extractor \textit{V} and a classifier \textit{G}.
The feature extractor is pretrained on face recognition dataset (VGG) to serve as initializer for first layers of V. After features are extracted, the mean of pseudo-negative class $\mu{*}$ is calculated based on the average of current $\mu\textsubscript{new}$ and previous batch $\mu\textsubscript{old}$ of genuine samples features. They model the pseud-negative distribution as a Gaussian distribution using calculated $\mu\textsubscript{*}$ and covariance matrix $\sigma$ and sample pseudo-negative data from this distribution. Finally, the normal class features and sampled pseudo-negative features are concatenated and given to fully-connected layers for classification. The training procedure is supervised using cross-entropy loss over the output of classifier, and a pairwise confusion~\cite{dubey2018pairwise} over extracted features of network \textit{V} to eliminate identity-related information and improve overall performance.

\subsection{Generative Models}
Unlike discriminative models which tries to find the boundary between different classes, generative models estimate the distribution of each category. In recent FAS works, generative models have come to attention, especially to design more robust FAS structures. Specifically, generative models usually are used to extract spoof traces which have two benefits. First, spoof traces can be used solely or combined with other network’s output to determine input’s liveness. Second, spoof images can be directly synthesized by pixel-wise addition of spoof traces with live samples. 

A proper example of such method is conducted by jourabloo et al.~\cite{jourabloo2018face}, their approach is motivated by image de-nosing and de-blurring in which the goal of these networks is to output an uncorrupted image when a corrupted one is given. To this end, they propose a network including three sub-networks: De-Spoofing Net (DS Net), Visual Quality Net (VQ Net), and Discriminative Quality Net (DQ Net). The DS Net is an encoder-decoder network that produces the noise for the given input, where noise for a live frame is a zero-value frame. Additionally, DS output a 0/1 map in its bottleneck, and a zero-map indicates a live face while a one-map represents a spoof face. The VQ Net helps to make the reconstructed real image (Input image-noise) to be visually realistic; the DQ Net outputs a depth map for the reconstructed real image, which further improves noise modeling. To distinguish real and fake faces, three metrics are available: depth map, noise map, and 0/1 map, which functions as an additional loss and is implemented on the DS Net encoder’s output. They achieve the best classification performance via fusing depth and noise maps.

Although previous work was only limited to spoof traces in print and replay attacks, recent work focus to cover a wider range of attack types~\cite{liu2020disentangling}. Liu et al. use spoof trace estimation in a face image to detect PA and synthesize new spoof images by merging those traces with live faces. They suggest a generator that partitions spoof traces into global, low-level, and high-level traces that are used together to generate the final spoof trace. Additionally, since the generator is fed with both live and spoof samples, they explore an early spoof regressor in the generator to produce a one/zero map which helps the generator's discriminability. They also apply a multiscale discriminator for three resolutions (256, 128, and 64) to capture different spoof cues and compete with the generator. Moreover, since the ground truth spoof traces exist for the synthesized spoof images, they apply pixel-wise loss in the generator output in the extra supervision step. Finally, the decision is made upon the aggregation of the spoof traces and the early spoof regressor map.

Unlike aforementioned works for the generative approach, instead of teaching the model to estimate 2D spoof patterns, Zhang et al. \cite{zhang2020face} attempt to obtain the spoof patterns from high-dimensional representations. The disentangled learning is leveraged to decompose the latent space of facial images into two sub-space: liveness space and content space. Liveness features corresponds to the liveness-related information, while content features encodes liveness-unrelated information of the input images, such as ID and lighting. 
This process of feature decomposition helps to create synthetic real and spoof images as well,but there is no ground truth for generated spoof images to guarantee their quality. To solve this issue, the Local Binary Pattern (LBP) as texture hint and pseudo-depth are used as auxiliary information. These maps (Depth and LBP) are considered zero for spoof images.
On the other hand, there are two identical discriminators with varied input resolution; higher resolution guides the disentanglement net to generate finer details and a smaller input scale guides the disentanglement net to preserve more global information.
The final loss function is the weighted summation of image reconstruction loss, latent reconstruction loss, LBP net, and depth net.

\subsection{Neural Architectural Search}
In deep learning, the process of finding the optimal structure of the network requires numerous trials and errors, and even then, you may end up with a sub-optimal solution. If we can automate this process, we could achieve efficient structures for the problem. In neural architectural search (NAS), finding optimal design occurs by specifying  network space, cell space, and operation space. However, it should be noted that prior knowledge of the task at hand plays a crucial role in selecting suitable search elements. For the FAS task, numerous networks have been suggested, and still, in some scenarios, we are far away from industrial standards. In few recent works of FAS, we can see the presence of NAS, which have achieved remarkable results.

Yu et al.~\cite{yu2020searching} propose the FAS network named CDCN and extend it by suggesting a more complex structure called CDCN++. CDCN is very similar to DepthNet~\cite{liu2018learning} that we introduced in the Auxiliary task category, which is an FCN estimating the facial depth but with a nuance to the original DepthNet. Moreover, since in many traditional works of FAS, LBP has been used extensively to extract local gradient information, they introduce Central Difference Convolution, which indirectly encapsulates LPB advantages in convolution operation. Convolution operation has two steps, one for sampling and the other for aggregating. The sampling step is same as vanilla convolution, but in the aggregation step, pixel's value located in the center of kernel is subtracted from all pixel values in the receptive field. Then, to combine intensity level and gradient level information, they merge vanilla and CDC convolutions, and a hyperparameter controls its CDC part contribution; they refer to this generalized convolution for simplicity also CDC. Therefore, the CDCN is DepthNet with the new convolution and uses contrastive depth plus mean square error as its loss functions. 

As an improvement to CDCN, the NAS and a Multiscale Attention Fusion Module (MAFM) are added on top of the CDC in CDCN++. Two first layers of CDCN++ are called the stem, the two last layers are called the head; there are three cells at three levels (low-level, mid-level, and high-level) between these layers. Each cell contains six nodes, including an input node, four intermediate nodes, and an output node. Search space includes eight possible options for intermediate nodes, and the output node is a weighted summation of intermediate nodes. All intermediate layer enter the MAFM, which applies spatial attention with a varying kernel size (high-level/semantic features with small attention kernel size and low-level with large kernel size) to refine multilevel features, and the concatenated output is given to head layers to predict the final facial depth.

In addition to ~\cite{yu2020searching}, Yu et al. introduce an extended version of this work and contribute further improvements for it ~\cite{yu2020fas}. Same as CDC in their previous work, they suggest Central Difference Pooling (CDP): a vanilla average pooling that in the aggregation step, the central pixel's value is subtracted from all pixels' values in the receptive field. The second contribution of this work is the introduction of static-dynamic representation. The learning process of dynamic representation is a convex optimization problem using RankSVM, and during this process, temporal information of k consecutive frames is embedded into a single frame. Since some detailed appearance clues are lost in the final dynamic image, they create a static-dynamic representation by combining the dynamic image with static image (input frame) and applying min-max normalization.
In additions to mentioned enhancements, to find the best NAS strategy and search space, they conduct extensive experiments to find the optimal solution in three main steps. First, they consider a baseline by comparing two input types (static and static-dynamic), two operation spaces (vanilla, CDC), and two loss functions (cross-entropy and pixel-wise binary). In the subsequent stage, because of pixel-wise superiority, they remove CE loss from search options, and they move closer to FAS by having a network space like the one used in CDCN++. This new search space also considers spatial attention and CDP effects. Finally, they present domain/type aware NAS, which exploits a meta learner to cover performance drop for unseen data or attack types.

\section{Datasets}\label{sec:datasets}
The availability of public datasets plays a vital role in advancement of new FAS technologies and reproducing the reported results. In this section, we provide detailed information on major publicly available FAS dataset (Table \ref{tab:fas-datasets}) and their production process. Further, we note the advantages or shortcomings of each dataset during our study. The benefits of this are twofold:
\begin{enumerate}
\item Envisions researchers to exploit available datasets effectively based on the datasets' attributes and their FAS system requirements.
\item It guides researchers to generate more valuable datasets to ease further FAS developments.
\end{enumerate}

\paratitle{NUAA Impostor.} NUAA Impostor~\cite{tan2010face} is gathered in three sessions with 15 unique subjects in which not all subjects participate in all sessions. Each video is recorded for 25 seconds at 20 fps, and some frames are selected for the dataset. The dataset is divided into train (sessions 1 and 2) and test sets (session 3). There are nine subjects in the train set and contains 1,743 real and 1,748 spoof images (3,497 in total). In the test set, there exist 3,362 real images from other six subjects and some subjects from the train set, and 5,761 spoof images from all 15 subjects (9,123 images in total).
For real data generation, participants are asked to look frontally at a web camera ($640\times480$ resolution) with a neutral face avoiding any facial or body movement. For spoof attacks, high-quality photos of subjects are printed on photographic papers in two sizes ($6.8\times10.2$ cm (small) and $8.9\times12.7$ cm (big)), and on 70g A4 papers using an unspecified HP color printer. These samples are then presented in front of the camera to create spoof attacks. This is the first publicly available FAS dataset that paved the path for following rigorous investigations. 

\paratitle{Yale-Recaptured.} Yale-Recaptured~\cite{peixoto2011face} uses Yale Face dataset to create a FAS dataset by adding photo replay attacks from existing live images. The dataset consists of 640 real faces and 1,920 photo attacks. Spoof images are taken at 50cm distance from the LCD screen and are cropped to center the face images. Finally, the images are converted to grayscale and resized to $64\times64$ pixels. For dataset creation, they cover three displays: an LG Flatron L196WTQ Wide 1900, a CTL 171Lx 1700 TFT, and a DELL Inspiron 1545 notebook; following two cameras are used as acquisition devices: 
a Kodak C813 with 8.2 megapixels and a Samsung Omnia i900, with 5 megapixels. This dataset is the second FAS dataset after the NUAA Imposter dataset and the first FAS dataset that comprised photo replay attack.

\paratitle{The Replay-Attack Family.}
The Replay-Attack dataset~\cite{chingovska2012effectiveness} and its subsets (the Print-Attack dataset~\cite{anjos2011counter} and the Photo-Attack dataset~\cite{anjos2014motion}) are made of 15s real videos and 10s spoof samples of 50 different subjects. The dataset contains 1,000 spoof videos and 300 real videos (100 videos for face verification and other 200 videos for FAS) and includes three common attack types: print attack, photo attack, and video attack. Real videos, for each subject are recorded three times (Apple 13-inch MacBook laptop at 25 fps with $320\times240$ resolution) at two different stationary conditions:
\begin{itemize}
\item Controlled: Background is uniform, and a fluorescent lamp illuminates the scene.
\item Adverse: Background is nonuniform, and daylight illuminates the scene.
\end{itemize}
In addition, extra data (two photographs and two video clips) is gathered for every person in each stationary condition for spoof attacks. The first photograph/video clip was recorded using iPhone 3 GS (3.1-megapixel camera) and the second using a high-resolution 12.1-megapixel Canon PowerShot SX200 IS camera. The border of the display media is not visible in the final spoof video, and the acquisition device for spoof attacks is the same as used for real videos. Furthermore, each spoof attack is recorded in two different attack sencarios: hand-based attacks, i.e, the attack media is held in hand, and fixed-support attacks, i.e, attack media is set on a fixed support.

iPhone and canon camera samples are displayed respectively on iPhone sceen (with $480\times320$ resolution) and iPad 1 screen (with $1024\times768$ resolution).
For print attacks, high-resolution digital photographs are printed on plain A4 paper using a Triumph-Adler DCC 2520 color laser printer. In total, there are four print attacks per subject, two for each illumination condition. Hard copies of high-resolution digital photographs are printed on plain A4 papers to generate print attacks.The dataset is categorized into six protocols, and each one is labeled with a specific condition, type of attack, the device used to perform the attack, or different types of support to hold the device. Each anti-spoofing protocol in the dataset contains 200 videos of real videos plus various types of attacks.

This collection greatly improves FAS dataset creation since the earlier datasets contain very few subjects and only cover one attack type (print attack or photo replay attack). Additionally, it adds video replay attacks to existing spoof attacks and provides videos instead of images which are more valuable since they let researchers use temporal cues in their FAS methods. Therefore, prior datasets to the Replay-attack family rarely appear in recent works.

\paratitle{CASIA-MFSD.} 
CASIA-MFSD~\cite{zhang2012face} includes 50 subjects with 600 videos (450 spoof and 150 genuine) and covers three attacks (two print attacks and one video replay attack). All videos are captured in natural scenes, and subjects are asked to blink during recording. Since video quality is a determining factor in the FAS task in this dataset, the data is gathered in three qualities:
low quality (unspecified long-time used USB camera with 640×480 resolution), normal quality (unspecified usb camera with 640x480 resolution), and high quality (Sony NEX-5 camera with a maximum resolution of 1920×1080, but only face region with $1280\times720$ is stored, which is more efficient to work with as the final video).

For wrapped photo attack, $1920\times1080$ images of Sony NEX-5 camera are printed on copper papers to have higher resolution than A4 paper, and the same photos in the previous step are used for the cut photo attacks. To simulate blink behavior, the eyes' regions are cut off, and an attacker hides behind and exhibits blinking through the holes, or one intact photo is tightly placed behind the cut photo and moved to mimic blinking. For video attack generation, genuine videos with $1280\times720$ resolution from the Sony NEX-5 camera are displayed on an iPad. However, the actual video resolution is downgraded to iPad resolution.

This dataset lacks photo attacks but includes other essential variations. Such variations are advanced factors like quality and print attack alternatives, such as wrapped paper and cut paper, which are more challenging attack types. In addition, since this dataset is categorized by different qualities, it can be used to benchmark the impact of quality on  FAS approaches.

\paratitle{3DMAD.} 
3DMAD~\cites{erdogmus2014spoofing} aims at one specific type of attack named 3d mask attacks, and includes 170 real videos and 85 mask attacks of 17 subjects. For 3d mask production, one frontal and two profile images of each subject are used on ThatsMyFace.com to create life-size wearable masks and paper-cut masks. The wearable masks are made out of a hard resin composite in full 24-bit color with holes at the eyes and the nostrils. To create spoof attacks, Microsoft Kinect for Xbox 360 is used to get RGB (8-bit) and depth data (11-bit) with  $640\times480$ resolution at 30 fps. The videos are collected in three different sessions in which the background of the scene is uniform, and the lighting is adjusted to minimize the shadows casted on the face.  A single operator performs first and second sessions for real videos (two weeks apart) and the third session for mask attacks. For each person, five videos of 10 seconds length are captured at each session. Afterward, eye positions for each video are annotated manually at every 60th frame and linearly interpolated for the rest. Further, the uploaded images and the paper-cut masks are accessible in the dataset. 3DMAD is the first FAS dataset that provides depth data and introduces 3d mask attacks which enables researchers to counter such attack type. However, there are few subjects, a limited amount of data, and only one attack type.

\paratitle{MSU-MFSD.} 
The MSU-MFSD~\cite{wen2015face} dataset consists of 110 genuine videos and 330 video clips of photo and video attacks of 55 subjects. For both real and spoof videos, the following acquisition devices are used: built-in camera in MacBook Air 13 with $640\times480$ resolution, and the front camera in the Google Nexus 5 with $720\times480$ resolution. The average frame rate is about 30 fps, and the duration of each video is at least 9 seconds (the average duration is $\approx12$ seconds). 
For real videos, subjects have $\approx50$cm distance to the camera. 

For replay attacks, following settings are utilized for capturing real and spoof videos:
\begin{enumerate}
\item Canon 550D Single-lens reflex (SLR) camera ($1920\times1088$) for recording and iPad Air screen ($2048\times1536$) for displaying at $\approx20$cm distance from the camera.
\item iPhone 5S rear camera ($1920\times1080$) for recording and its own screen ($1136\times640$) for displaying at $\approx10cm$ distance from the camera.
\end{enumerate}

For print attack: pictures are captured from each subject using the Canon 550D camera ($5184\times3456$) and printed on an A3 paper ($11.7\times16.5$) using an HP Color Laserjet CP6015xh printer ($1200\times600$ dpi). The average distance from the camera for the printed photo attack is $\approx40cm$.

MSU-MFSD is the first dataset that employs a mobile phone camera as an acquisition device, making gathered data more similar to FAS in real-world cases. Moreover, it has higher quality print attacks than prior works since high resolution ($5184\times3456$) images are printed in large size (A3 paper) using a high-quality color printer.

\paratitle{HKBU-MARs V2.} 
HKBU-MARs V2~\cite{liu20163d} contains 1,008 videos of 3D mask attacks and genuine videos (504 each type) of 12 subjects and 12 masks from two companies (half from ThatsMyFace and half are from Real-F).  The dataset has vast variation in terms of illumination conditions and acquisition devices. 
There are seven different cameras, including three stationary ones: a Logitech C920 web camera ($1280\times720$), an economic class industrial camera ($800\times600$), and Canon EOS M3 camera ($1280\times720$), plus three smartphones (Nexus 5, iPhone 6, Samsung S7), and Sony Tablet S. The illumination conditions to cover the typical scenes are as follow:
Room light (the most common condition) in which it's the office light,  low light, bright light and warm light (the typical variations of the in-door light), sidelight, and upside light to simulate the possible harsh lighting condition.

For each subject, 42 videos (each around 10 seconds) are recorded as the combination of seven cameras and six lighting variations at 30 fps except EOS M3 (50 fps) and industry camera (20 fps). They use a tripod for stationary cameras at around 80 cm distance from the camera, and for smartphones, the subject holds the phone in hand at a comfortable distance.

Compared to 3DMAD (prior 3D mask dataset), it has five fewer subjects but covers two types of masks, a wide range of acquisition devices with various qualites, and diverse illumination conditions, which make all valuable data for finding robust FAS method against 3D mask attacks. 

\paratitle{MSU-USSA.} 
MSU-USSA~\cite{patel2016secure} contains 1,140 authentic images (with an average resolution of $705\times865$) and 9,120 spoof attacks (print and replay attacks) of 1,140 subjects. For dataset creation, web faces dataset collected in~\cite{wang2013retrieval} is used which have 1,000 subjects, and contains images of celebrities taken under various backgrounds, illumination conditions,and resolutions. However, images from previous step are filtered to retain only a single frontal face image. Moreover, they add other 140 subjects as follows: 50 from the Idiap, 50 from CASIA-FASD, and 40 from the MSU-MFSD. 

\textbf{Photo replay attacks:} there are 6,840 images created by using both the front ($1280\times960$) and rear ($3264\times2448$) cameras of Google Nexus 5 and display devices such as MacBook ($2880\times1080$), Nexus 5 ($1920\times1080$), and Tablet ($1920\time1200$) screens. During replay attack generation, has been tried to minimize the illumination reflections.

\textbf{Print attack:} includes 2,280 images generated in the following process. Subject's images are first scaled to cover of a matte ($8.5\times11$ inch white paper) as much as possible while maintaining the original image aspect ratio to minimize distortions, and then HP Color Laserjet is used for printing. Additionally, they placed the photos in a manner to minimize reflection from ambient lighting inside their laboratory, and both the frontal and rear cameras of Nexus 5 were used to capture print attacks. 

There are way more subjects in this dataset than any preceding one, and since it comes from different datasets, it also has great diversity in background, lighting conditions, and image quality which are all valuable. It also uses multiple display devices for spoof medium, which brings more diversity. However, it does not cover video replay attacks, plus the dataset format is in images instead of video, which does not provide temporal information. 

\paratitle{Replay-mobile.} 
Replay-mobile~\cite{costa2016replay} contains 1,190 video clips (390 real, 640 spoof, and 160 enrollment) of photo and video attack attempts by 40 clients. The dataset has been collected in two sessions, separated by an interval of two weeks.
In the first session, both enrollment videos (videos for face verification) and media for manufacturing the attacks were collected under two different illumination conditions, namely light-on (electric lights in the room are switched on) and light-off (electric lights are turned off). For attack generation, 10-second videos were recorded via the rear camera of the LG-G4 smartphone ($1920\times1080$), and 18 Mega pixel Nikon Coolpix P520 was used to capture images. The scene's background is homogeneous in both scenarios, and a tripod is used for the capturing device.

In the second session, each client recorded ten videos (about 10 seconds at 30fps with $720\times1280$ resolution), using an iPad Mini 2 tablet and another using an LG-G47 smartphone under the following five different scenarios:
\begin{enumerate}
\item \textbf{Controlled:} The background of the scene is uniform, the light in the office is switched on, and the window blinds are down.
\item \textbf{Adverse:} The background of the scene is uniform, the light in the office is switched off, and the window blinds are halfway up.
\item \textbf{Direct:} The background of the scene is complex, and the user is facing a window with direct sunlight while capturing the video.
\item \textbf{Lateral:} The scene's background is complex, and the user is near a window and receives lateral sunlight while capturing the video.
\item \textbf{Diffuse:} The video is captured in an open hall with a complex background and diffuse illumination.
\end{enumerate}

In this session, the user is asked to stand, hold the mobile device at eye level, and center the face on the screen of the video capture application. A Philips 227ELH monitor with 1920×1080 resolution is used as a display device for replay attacks; for print attacks, photos are printed on A4 matte paper (using a Konica Minolta ineo+ 224e color laser printer).
Each attack is recorded on each mobile device (tablet and smartphone) for 10 seconds, and for replay attacks, the acquisition device is supported on a fixed support. However, print attacks are carried out in two different attack modes: hand-held attack, where the operator holds the capture device, and fixed-support attack, where the capturing device is fixed.

Replay-mobile provides all common attack types and has additional enrollment videos for face verification. Videos have higher quality than MSU-MFSD, but the number of subjects is slightly fewer than recent datasets, and they only have one display device for replay attacks.

\paratitle{Oulu npu.} 
Oulu npu~\cite{boulkenafet2017oulu} includes 5,960 5-second videos (1,980 real and 3,960 spoof) corresponding to 55 subjects (15 female and 40 male). Six real videos are recorded via mobile phones’ front cameras for each subject, and the clients are asked to hold the mobile device like they are being authenticated. Real videos are gathered in three different sessions (two videos in each session) with one-week intervals and the following settings:
\begin{enumerate}
\item Session 1: An open plan with lights on and windows located behind the subject.
\item Session 2: A meeting room with only electronic light as illumination.
\item Session 3: A small office with lights on and windows located in front of the subject.
\end{enumerate}

Six smartphones with high-quality front cameras in the price range from €250 to €600 are covered for the data collection:
\begin{itemize}
\item Samsung Galaxy S6 edge with 5 MP frontal camera.
\item HTC Desire EYE with 13 MP frontal camera.
\item MEIZU X5 with 5 MP frontal camera.
\item ASUS Zenfone Selfie with 13 MP frontal camera.
\item Sony XPERIA C5 Ultra Dual with 13 MP frontal camera.
\item OPPO N3 (Phone 6) with 16 MP rotating camera
\end{itemize}

Real videos are recorded using mobile phones’ front cameras with the same camera software on all devices and $1920\times1080$ resolution. Besides, an extra 16 MP photo and Full HD video are captured using the rear camera of Samsung Galaxy S6 to create spoof attacks. Then for print attacks, the high-resolution images are printed on A3 glossy paper using Canon imagePRESS C6011 and Canon PIXMA iX6550 printers. And for video replay attacks, the high-resolution videos are replayed on 19 inch Dell UltraSharp 1905FP display with $1280\times1024$ resolution and an early 2015 Macbook 13” laptop with a Retina display of $2560\times1600$ resolution.

The print and video replay attacks are then recorded using the frontal cameras of the six mobile phones. For print attacks, the prints are held by the operator and captured with stationary capturing devices. In contrast, when recording the video replay attacks, both the capturing devices and PAIs are stationary. Moreover, the background scene of the attacks are matched to the real videos during each session, plus the attack videos do not contain the bezels of the screens or edges of the prints. The subsequent protocols for various experiments are designed:
\begin{enumerate}
\item Protocol I: The first protocol evaluates the generalization of the face PAD methods under different environmental conditions, namely illumination and background scene; hence they separate session 3 for the test. 
\item Protocol II:  The second protocol evaluates the effect of the PAI variation on the performance of the face PAD methods by using different displays and printers.
\item Protocol III: To evaluate the acquisition device effect, the Leave One Camera Out (LOCO) protocol is designed in which, in each iteration, the real and the attack videos recorded with five smartphones are used for training and evaluation and test conducted by the remaining mobile phone.
\item Protocol IV: the previous three protocols are combined to simulate the real-world operational conditions; the generalization ability of FAS methods are evaluated simultaneously across previously unseen illumination conditions, background scenes, PAIs, and input sensors.
\end{enumerate}
The details of all protocols can be seen in (Table \ref{tab:oulu-protocols}).

\begin{table}[h]
\centering
\caption{\label{tab:oulu-protocols} The detailed information about the video recordings in the train, development and test sets of each protocol.}
\resizebox{\textwidth}{!}{%
\begin{tabular}{|c|c|c|c|c|c|c|c|c|}
\hline
Protocol                      & Subset & Session       & Phones   & Users & Attacks created using    & \# real videos & \# attack videos & \# all videos \\ \hline
\multirow{3}{*}{Protocol I}   & Train  & Session 1,2   & 6 Phones & 1-20  & Printer 1,2; Display 1,2 & 240            & 960              & 1,200          \\ \cline{2-9} 
 & Dev  & Session 1,2   & 6 Phones & 21-35 & Printer 1,2; Display 1,2 & 180 & 720 & 900  \\ \cline{2-9} 
 & Test & Session 3     & 6 Phones & 36-55 & Printer 1,2; Display 1,2 & 240 & 960 & 1,200 \\ \hline
\multirow{3}{*}{Protocol II}  & Train  & Session 1,2,3 & 6 Phones & 1-20  & Printer 1; Display 1     & 360            & 720              & 1,080          \\ \cline{2-9} 
 & Dev  & Session 1,2,3 & 6 Phones & 21-35 & Printer 1; Display 1     & 270 & 540 & 810  \\ \cline{2-9} 
 & Test & Session 1,2,3 & 6 Phones & 36-55 & Printer 2; Display 2     & 360 & 720 & 1,080 \\ \hline
\multirow{3}{*}{Protocol III} & Train  & Session 1,2,3 & 5 Phones & 1-20  & Printer 1,2; Display 1,2 & 300            & 1,200             & 1,500          \\ \cline{2-9} 
 & Dev  & Session 1,2,3 & 5 Phones & 21-35 & Printer 1,2; Display 1,2 & 225 & 900 & 1,125 \\ \cline{2-9} 
 & Test & Session 1,2,3 & 1 Phones & 36-55 & Printer 1,2; Display 1,2 & 60  & 240 & 300  \\ \hline
\multirow{3}{*}{Protocol VI}  & Train  & Session 1,2   & 5 Phones & 1-20  & Printer 1; Display 1     & 200            & 400              & 600           \\ \cline{2-9} 
 & Dev  & Session 1,2   & 5 Phones & 21-35 & Printer 1; Display 1     & 150 & 300 & 450  \\ \cline{2-9} 
 & Test & Session 3     & 1 Phones & 36-55 & Printer 2; Display 2     & 20  & 40  & 60   \\ \hline
\end{tabular}%
}
\end{table}

Oulu npu uses six mobile phones as acquisition devices which are significantly more than prior datasets. Moreover, two printers are used for print attacks and have two display devices for video replay attacks. Besides, the videos are recorded in three sessions, bringing different illumination and environment scenes. Finally, by combining all these variables, valuable protocols were designed for insightful evaluations.

\paratitle{SiW.} 
SiW~\cite{liu2018learning} provides 4,620 live and spoof videos (8 live and 20 spoof videos for each subject) at 30 fps from 165 subjects. The live videos are captured with two high-quality cameras (Canon EOS T6, Logitech C920 webcam) in $1920\times1080$ resolution with different PIE variations in four sessions. In session 1, subjects move their heads at varying distances to the camera. In session 2, subjects changes the yaw angle of their head within [-90$^{\circ}$, 90$^{\circ}$] and makes different facial expressions. In sessions 3, 4, subjects repeat sessions 1, 2 while the collector moves the light source around the face from different orientations.

Two print and four video replay attacks are provided for each subject. To generate different qualities of print attacks, a high-resolution image ($5184\times3456$) is captured and a frontal-view frame from a live video is extracted for lower quality print attacks for each subject, and images are printed by HP color LaserJet M652 printer. And for replay attacks, they select four spoof mediums: Samsung Galaxy S8 ($1920\times1080$), iPhone 7 Plus ($1334\times750$), iPad Pro (2048x1536), and Asus MB168B ($1366\times768$) screens. They randomly select two of the four high-quality live videos to display in the spoof mediums for each subject.

SiW contains three times the subjects of Oulu-NPU (most number of subjects prior to this dataset) and covers diverse races. Additionally, subjects have different facial expressions, and face pose is not limited to frontal. Besides, four display devices are used for various replay attacks. However, only two acquisition devices are employed which were six in the Oulu npu dataset.

\paratitle{Rose-Youtu.} 
Rose-Youtu~\cite{li2018unsupervised} consists of 3,497 (899 real and 2,598 spoof) videos with 20 subjects. There are 150-200 video clips for each subject with an average duration of around 10 seconds. Five mobile phones were used to collect the dataset: (a) Hasee smartphone ($640\times480$), (b) Huawei smartphone ($640\times480$), (c) iPad 4 ($640\times480$), (d) iPhone 5s ($1280\times720$) and (e) ZTE smartphone ($1280\times720$). Front cameras capture all face videos, and the standoff distance between face and camera is about 30-50 cm.
For each client, there are 25 genuine videos (five devices with five scenes), which each scene is a different illumination condition in an office environment, and if the client wears eyeglasses, there will be another 25 videos. Three attack types are covered, including print attack, video replay attack, and mask attack. For print attacks, images with still printed paper and quivering printed paper (A4 size) are used, and for the video replay attacks, a face video is displayed on the Lenovo LCD screen ($4096\times2160$) and Mac screen ($2560\times1600$). For the mask attacks, masks with and without cropping are considered. Furthermore, the face videos are captured with different backgrounds, which guarantees the face videos are coupled with varying illumination conditions.

Rose-Youtu covers different environments, lighting variations, five acquisition sensors (more than average), and has two high-resolution display devices. However, it only has 25 subjects and does not include photo replay attack. Moreover, the capturing devices have a low resolution in comparison with previous datasets.

\paratitle{SiW-M.} 
SiW-M~\cite{liu2019deep} includes 1,630 videos (660 real and 968 spoof) with 5-7 sec duration from 493 subjects. 13 types of spoof attacks and two spoofing scenarios are considered: impersonation, in which attacker tries to be recognized as someone else, and in obfuscation, attacker aim at hiding his identity. 
For all five mask attacks, three partial attacks, obfuscation makeup, and cosmetic makeup, 1080P HD videos are recorded using a Logitech C920 webcam and a Canon EOS T6 camera. For impersonation makeup, 720P videos are collected from Youtube due to the lack of special makeup artists. For print and replay attacks, videos that an off-the-shelf face anti-spoofing algorithm~\cite{liu2018learning} fails (predicts live) are gathered.
Subjects are diverse in ethnicity and age, and live videos are collected in 3 sessions:
\begin{enumerate}
\item Room environment where subjects are recorded with few variations such as pose, lighting, and expression (PIE).
\item Different and much larger room where subjects are also recorded with PIE variations.
\item Mobile phone mode in which the subjects are moving while the phone camera is recording, so that extreme pose angles and lighting conditions are introduced.
\end{enumerate}
Similar to print and replay videos, the same FAS algorithm\cite{liu2018learning} is used to find out the videos where it fails (predicts spoof).

SiW-M has great number of subjects, wide range of spoof types and covers various environments and other effecting factors. However, it only employs two acquisition devices, and no information is available about display device(s). Additionally, there are only 217 videos for print and replay attacks (fewer than 25\% of all spoof videos), which are considered as major attack types.

\paratitle{CelebA-Spoof.} 
CelebA-Spoof~\cite{zhang2020celeba} provides 625,537 pictures from 10,177 subjects (202,599 real and 422,938 spoof images), with each subject having 5 to 40 authentic images. The live data are directly inherited from the CelebA dataset, and it covers face images with large pose variations and background clutters.
When creating spoof attacks from live images, for each subject with more than k source images, images are ranked according to the face size with the bounding box provided by CelebA and Top-k ones are selected. k is set at 20, and as a result, 87,926 source images are selected from 202,599. This dataset covers following attack types:print attack, paper cut attack, 3D mask attack, and photo attack. To improve the generalization and diversity of the dataset, three factors are leveraged (angle, attack medium, and acquisition device): 
\begin{enumerate}
\item Five Angles: all spoof types need to traverse all five types of angles, including vertical, down, up, forward, and backward. The angle of inclination is between [$-30^{\circ}$, $30^{\circ}$]. 
\item Four Shapes: there are a total of four shapes, i.e., normal, inside, outside, and corner. 
\item Four types of acquisition devices: 24 acquisition devices in four types are employed, i.e., PC, camera, tablet, and phone. These devices are equipped with different resolutions, ranging from 40 million to 12 million pixels which can be categorized into three quality groups (low quality, middle quality, and high quality) .
\end{enumerate}

This dataset is the most extensive dataset in terms of number of subjects and data quantity. Plus, it covers various spoof attacks with different qualities and PIEs and has 24 acquisition devices which bring more diversity than any preceding datasets. The lack of temporal information and limitation to static attacks are the only drawbacks for this dataset. Besides, it is worth noting that like MSU-USSA\cite{patel2016secure} no real subject is present during dataset creation.

\paratitle{RECOD-MPAD.} RECOD-MPAD~\cite{ almeida2020detecting} includes 45 subjects (30 men and 15 women ranging from 18 to 50 years old) containing around 144,000 frames (28,800 genuine and 115,200 spoof) and covers two print attacks and two photo replay attacks. The genuine videos ($\approx10$ seconds) are recorded via frontal camera of MOTO G5 smartphone and MOTO X Style XT1572 in following sessions to encapsulate various lighting and environment conditions:
\begin{enumerate}
\item Session 1: Outdoors, direct sunlight on a sunny day. 
\item Session 2: Outdoors, in a shadow (diffuse lighting). 
\item Session 3: Indoors, artificial top light. 
\item Session 4: Indoors, natural lateral light (window or door). 
\item Session 5: Indoors, lights off (noisy).
\end{enumerate}
Users were asked to hold the phone as if they were using frontally and slowly rotate their heads to change lighting and background, which causes variation within frames of a video. To create photo replay attacks they used unspecified 42-inch CCE TV and unspecified 17-inch HP monitor as their display devices. For print attacks, two frames from genuine videos were extracted and printed on A4 paper using unspecified printer; first printed photo was recaptured indoors, and the second one outdoors with more light. Finally, 64 equally-spaced frames were extracted from each video and for those frame that DLib failed to detect eye centers, they manually annotated and stored. The dataset contains 24 users with 77,000 frames in training set, 6 users with 19,200 frames in validation set, and 15 users with 48,000 frames in test set. 
Subjects stand almost still in genuine videos, but there are variations among different frames of a single video due to slight head rotation, which makes frames more informative. The number of subjects and dataset’s size is acceptable since it is comparable with most recent datasets. However, same as CelebA-Spoof~\cite{zhang2020celeba} it only includes static attacks and since dataset is in image format it lacks temporal information of a video. 

\paratitle{CASIA-SURF 3DMASK} CASIA-SURF 3DMASK~\cite{yu2020fas} includes 1,152 videos (288 genuine and 846 3D mask attacks) of 48 subjects (21 male and 27 females), and six lighting conditions are considered to mimic real-world environment: normal, back-light, front-light, side-light, outdoors in shadow and outdoors in sunlight. The videos are recorded with $1920\times1080$ resolution at 30fps using smartphones in several brands (i.e., Apple, Huawei and Samsung). While previous 3D mask datasets ~\cite{liu20163d,erdogmus2014spoofing} only consider masks with no decorations, in CASIA-SURF 3DMASK two realistic decorations (i.e., masks with/without hair and glasses) are covered. Therefore attacks are created via three masks, six lighting conditions for each subject. CASIA-SURF 3DMASK contains more subjects and data than prior 3D mask datasets, Moreover, it is the first 3D mask dataset that includes realistic decorations on masks. 

\paratitle{CASIA-SURF HiFiMask (HiFiMask).} HiFiMask~\cite{liu2021contrastive} includes 54,600 videos (13,650 genuine and 40,950 3D mask attacks) from 75 subjects (25 subjects in three skin tones: yellow, white, and black), and utilize three masks in various materials (i.e., plaster, resin, and transparent). They are 225 masks in total, and videos are recorded in six different scenes (i.e., white light, green light, periodic three color light, outdoor sunshine, outdoor shadow, and motion blur), with six directional lighting conditions (i.e., normal light, dim light, bright light, black light, side light, and top light). Videos are collected using 7 mainstream capturing devices (i.e., iPhone11, iPhoneX, MI10, P40, S20, Vivo, and HJIM).

HiFiMask is a challenging dataset due to high-fidelity 3D masks and temporal light interference; thus, texture-based methods and rPPG-based methods fail to perform well against HiFiMask. It is the latest 3D mask dataset which contains the most number of subjects, videos, and the most diverse combination of scenes and lighting conditions. Additionally, it is the first 3D mask dataset that employs mask in different material and considers skin tone as a differentiating factor. 

\FloatBarrier
\begin{landscape}
\tiny
\begin{longtable}[c]{|c|c|c|c|c|c|c|c|}
\caption{Face Anti-spoofing datasets.}
\label{tab:fas-datasets}\\
\hline
dataset &
  \begin{tabular}[c]{@{}c@{}}Year of\\ release\end{tabular} &
  \begin{tabular}[c]{@{}c@{}}\# Data (V/I)\\ (Live, Spoof)\end{tabular} &
  \# Subj. &
  Attack type &
  Acquisition devices &
  Spoof medium &
  Printers \\ \hline
\endhead
NUAA &
  2010 &
  (5,105, 7,509)(I) &
  15 &
  Print attack &
  Webcam ($640\times480$) &
  \begin{tabular}[c]{@{}c@{}}A4 paper\\ photographic papers:\\ 6.8 × 10.2 cm (small)\\ 8.9 × 12.7 cm (big)\end{tabular} &
  Unspecified HP color printer \\ \hline
Yale-Recaptured &
  2011 &
  (640, 1,920) (I) &
  10 &
  Photo attack &
  \begin{tabular}[c]{@{}c@{}}Kodak C813 (8.2MP),\\ Samsung Omnia i900 (5 MP)\end{tabular} &
  \begin{tabular}[c]{@{}c@{}}LG Flatron L196WTQ wide 1900\\ CTL 171Lx 1700 TFT\\ Dell Inspiron 1545 notebook\end{tabular} &
  No printer is used. \\ \hline
Replay-attack Family &
  \begin{tabular}[c]{@{}c@{}}2011\\ 2012\\ 2014\end{tabular} &
  (200, 1,000) (V) &
  50 &
  \begin{tabular}[c]{@{}c@{}}Print attack\\ photo attack\\ Video replay attack\end{tabular} &
  \begin{tabular}[c]{@{}c@{}}Apple 13-inch Mackbook laptop ($320\times240$),\\ iPhone 3GS (3.1MP),\\ Canon PowerShot SX200 IS (12.1MP)\end{tabular} &
  \begin{tabular}[c]{@{}c@{}}iPhone 3GS screen ($480\times320$)\\ iPad screen ($1024\times768$)\\ A4 paper\end{tabular} &
  \begin{tabular}[c]{@{}c@{}}Triumph-Adler DCC 2520\\ color laser printer\end{tabular} \\ \hline
CASIA-MFSD &
  2012 &
  (150, 150) (V) &
  50 &
  \begin{tabular}[c]{@{}c@{}}Warped photo attack\\ Cut photo attack\\ Video attack\end{tabular} &
  \begin{tabular}[c]{@{}c@{}}Low-quality camera ($640\times480$),\\ Normal-quality camera ($480\times640$),\\ Sony NEX$-$5 ($1920\times1080$),\\ and face region ($1280\times720$)\end{tabular} &
  \begin{tabular}[c]{@{}c@{}}iPad\\ copper paper\end{tabular} &
  Not reported. \\ \hline
3DMAD &
  2014 &
  (170, 85) &
  17 &
  3D Mask attack &
  Microsoft Kinect for Xbox 360 ($640\times480$) &
  \begin{tabular}[c]{@{}c@{}}Masks provided from\\  ThatsMyFace: hard resin composite\\  in full 24-bit color with holes\\  at the eyes and the nostrils\end{tabular} &
  No printer is used. \\ \hline
MSU-MFSD &
  2015 &
  (110, 330) (V) &
  55 &
  \begin{tabular}[c]{@{}c@{}}print attack\\ Video replay attack\end{tabular} &
  \begin{tabular}[c]{@{}c@{}}MacBook air 13-inch ($640\times480$),\\ Google nexus 5 front camera ($720\times480$)\\ Only for spoof generation:\\ Canon 550D Single-lens reflex \\ (SLR) ($1920\times1080$),\\ iPhone 5S rear camera ($1920\times1080$),\\ for print attack images($5184\times3456$)\end{tabular} &
  \begin{tabular}[c]{@{}c@{}}iPad Air screen ($2048\times1536$)\\ iPhone 5S rear camera ($1920\times1080$)\\ A3 paper\end{tabular} &
  \begin{tabular}[c]{@{}c@{}}HP Color Laserjet CP6015xh\\ printer ($1200\times600$ dpi)\end{tabular} \\ \hline
HKBU-MARs V2 &
  2016 &
  (504, 504) (V) &
  12 &
  3D mask attack &
  \begin{tabular}[c]{@{}c@{}}Logitech C920 web camera ($1280\times720$),\\ economic class industrial camera ($800\times600$),\\ Canon EOS M3 ($1280\times720$),\\ Nexus 5, iPhone 6, Samsung S7 Sony Tablet S\end{tabular} &
  \begin{tabular}[c]{@{}c@{}}6 masks from ThatsMyFace\\ 6 masks from Real-F\end{tabular} &
  No printer is used. \\ \hline
MSU-USSA &
  2016 &
  (1,140, 9,120)(I) &
  1140 &
  \begin{tabular}[c]{@{}c@{}}Photo attack \\ Print attack\end{tabular} &
  \begin{tabular}[c]{@{}c@{}}Google Nexus 5\\ front camera ($1280\times960$)\\ rear camera ($3264\times2448$)\end{tabular} &
  \begin{tabular}[c]{@{}c@{}}MacBook ($2880\times1800$)\\ Nexus 5 ($1920\times1080$)\\ Tablet ($1920\times1200$)\\  $8.5\times11$ inch paper \end{tabular} &
  HP Color Laserjet \\ \hline
Replay-mobile &
  2016 &
  \begin{tabular}[c]{@{}c@{}}(390, 640) (V)\\ 160 (enrollment)\end{tabular} &
  40 &
  \begin{tabular}[c]{@{}c@{}}Photo attack\\  Print attack\end{tabular} &
  \begin{tabular}[c]{@{}c@{}}iPad Mini 2 tablet,\\ LG-G47 smartphone both with $720\times1280$\\ Attack generation:\\ LG-G4 smartphone ($1920\times1080$),\\ Nikon Coolpix P520 (18 MP)\\ for capture image\end{tabular} &
  \begin{tabular}[c]{@{}c@{}}Philips 227ELH \\ monitor ($1920\times1080$),\\ A4 paper\end{tabular} &
  \begin{tabular}[c]{@{}c@{}}Konica Minolta ineo+ 224e\\ color laser printer\end{tabular} \\ \hline
Oulu npu &
  2017 &
  (1,980, 3,960) (v) &
  55 &
  \begin{tabular}[c]{@{}c@{}}Print attack\\ Video replay attack\end{tabular} &
  \begin{tabular}[c]{@{}c@{}}phone1: Samsung Galaxy S6 edge \\ frontal camera (5 MP),\\ phone2: HTC Desire EYE\\  frontal camera (13 MP),\\ phone3: MEIZU X5 \\ frontal camera (5 MP),\\ phone4: ASUS Zenfone Selfie \\ frontal camera (13 MP),\\ phone5$:$ Sony XPERIA C5 Ultra Dual \\ frontal camera (13 MP),\\ phone 6: OPPO N3 rotating camera (16MP)\\ All videos are recorded with\\ $1920\times 1080$ resolution\\ For spoof material generation:\\ Samsung Galaxy S6 rear camera (16 MP)\\ photo and Full HD video\end{tabular} &
  \begin{tabular}[c]{@{}c@{}}19-inch Dell UltraSharp\\ 1905FP display ($1280\times1024$)\\ Early 2015 Macbook 13-inch laptop\\ with a Retina display ($2560\times1600$)\\ A3 glossy paper\end{tabular} &
  \begin{tabular}[c]{@{}c@{}}Canon imagePRESS C6011,\\ Canon PIXMA iX6550\end{tabular} \\ \hline
SiW &
  2018 &
  (1,320, 3,300) (V) &
  165 &
  \begin{tabular}[c]{@{}c@{}}Print attack\\ Video replay attack\end{tabular} &
  \begin{tabular}[c]{@{}c@{}}Canon EOS T6\\ Logitech C920 webcam  ($1920\times1080$)\end{tabular} &
  \begin{tabular}[c]{@{}c@{}}Samsung Galaxy S8 ($1920\times1080$)\\ iPhone 7 plus ($1334\times750$)\\ iPad Pro ($2048\times1536$)\\ Asus MB168B ($1366\times768$)\end{tabular} &
  \begin{tabular}[c]{@{}c@{}}HP color\\ LaserJet M652 printer\end{tabular} \\ \hline
SiW-M &
  2019 &
  (660, 968) (V) &
  493 &
  \begin{tabular}[c]{@{}c@{}}3D mask attack\\ partial attack\\ obfuscation makeup\\ cosmetic makeup\\ impersonation makeup\\ print attack\\ replay attack\end{tabular} &
  \begin{tabular}[c]{@{}c@{}}Logitech C920 webcam ($1080\times1920$),\\ Canon EOS T6 camera ($1080\times1920$),\\ 720P Youtube videos\end{tabular} &
  Not reported. &
  Not reported. \\ \hline
CelebA-Spoof &
  2020 &
  (202,599, 422,938) (I) &
  10177 &
  \begin{tabular}[c]{@{}c@{}}Print attack\\ Paper cut attack\\ 3D mask attack\\ Photo attack\end{tabular} &
  \begin{tabular}[c]{@{}c@{}}24 sensors in 4 categories:\\ PC, Camera, Tablet, Phone\\ (12-40 MP)\end{tabular} &
  \begin{tabular}[c]{@{}c@{}}A4 paper\\ Poster\\ Photo\end{tabular} &
  Not reported. \\ \hline
  RECOD-MPAD &
  2020 &
  (28,800, 115,200) (I) &
  45 &
  \begin{tabular}[c]{@{}c@{}}Print attack\\ Photo replay attack\end{tabular} &
  \begin{tabular}[c]{@{}c@{}}MOTO G5\\ MOTO X Style XT1572\end{tabular} &
  \begin{tabular}[c]{@{}c@{}}A4 paper\\ unspecified 42-inch CCE TV\\ unspecified 17-inch HP monitor\end{tabular} &
  Not reported. \\ \hline
  CASIA-SURF 3DMASK &
  2020 &
  (288, 846) (V) &
  48 &
  \begin{tabular}[c]{@{}c@{}}3D mask attack\end{tabular} &
  \begin{tabular}[c]{@{}c@{}}smartphones in\\ Apple, Huawei and Samsung brands\end{tabular} &
  \begin{tabular}[c]{@{}c@{}}plaster\end{tabular} &
  No printer is used. \\ \hline
  HiFiMask &
  2021 &
  (13,650, 40,950) (V) &
  75 &
  \begin{tabular}[c]{@{}c@{}}3D mask attack\end{tabular} &
  \begin{tabular}[c]{@{}c@{}}iPhone11, iPhoneX, MI10\\ P40, S20, Vivo, and HJIM\end{tabular} &
  \begin{tabular}[c]{@{}c@{}}transparent, plaster, resin\end{tabular} &
  No printer is used. \\ \hline

\end{longtable}
\end{landscape}
\FloatBarrier

\newpage
\KOMAoptions{paper=portrait,pagesize}

\section{Results}\label{sec:results}
Accuracy can not properly represent the performance of biometric systems since error rate for different classes often have unequal significance or data from various classes are imbalanced. Hence, prevalent metrics are based on model's error for each class (live and spoof). Two major metrics to assess FAS systems are FAR (False Acceptance Rate) and FRR (False Rejection Rate) which are defined as follows:

\begin{align*}
  FAR = \frac{FP}{FP+TN}\\
  FRR = \frac{FN}{FN+TP}\\
\end{align*}

FP, TP, FN, and TN are respectively defined as False Positive (genuine access detected as attack), True Positive (Correctly detected spoof attack), False Negative (Spoof attack detected as genuine), and True Negative (genuine access not considered as attack).
Anjos et al.\cite{anjos2011counter} proposed HTER (Half Total Error Rate) which combines FAR and FRR in the subsequent manner:

\begin{align*}
  HTER = \frac{FAR+FRR}{2}\\
\end{align*}

It should be noted that model's threshold plays as a trade-off parameter for FAR and FRR; for HTER the threshold is selected at the EER point on the validation set, in which FAR equals FRR. After 2017, FAS models performance is reported based on the metrics defined in the standardized ISO/IEC 30107-3 metric, which was also suggested in Oulu-npu~\cite{boulkenafet2017oulu}. Two new metrics are Attack Presentation Classification Error Rate (APCER) which corresponds to FAR and Bona Fide Presentation Classification Error Rate (BPCER) which corresponds to FRR. For each new metric FRR and FAR is calculated for all PAIs and the maximum value is selected as APCER and BPCER. Similar to HTER, ACER is calclated as the average of APCER and BPCER at EER threshold on the validation sets.


\begin{align*}
  ACER = \frac{APCER+BPCER}{2}\\
\end{align*}

To evaluate FAS methods, commonly, the trained model on a training set is tested on its corresponding test set based on aforementioned evaluation metrics, which is known as intra-dataset evaluation. Evaluation is more valuable when there are numerous variations between training and test sets, so we consider Oulu-npu protocols for intra-dataset evaluation (Table~\ref{tab:oulu}), which as we explained in section~\ref{sec:datasets} in each protocol some effective factor is different between train and test sets. Recently, researchers are more concerned with FAS methods' performance when domain shift occurs because the same scenario applies when trained model in a lab goes to production in real-world applications. To better asses FAS methods for such cases, inter-dataset evaluations are designed in which test set comes from unseen dataset. When comparing performance of different FAS systems, we have provided two variations of inter-dataset evaluation: model is trained on a single dataset (Table~\ref{tab:inter-dataset2}), model is trained on multiple datasets (Table~\ref{tab:inter-dataset1})

\begin{table}[H]
\centering
\caption{The intra-testing results of four protocols of Oulu-NPU
dataset.}
\label{tab:oulu}
\resizebox{\textwidth}{!}{%
\begin{tabular}{|c|c|c|c|c|c|}
\hline
\multicolumn{1}{|l|}{Prot.} &
  \textit{Method} &
  Cues &
  APCER(\%) &
  BPCER(\%) &
  ACER(\%) \\ \hline
 &
  Auxiliary~\cite{liu2018learning} &
  Depth,   rPPg &
  1.6 &
  1.6 &
  1.6 \\ \cline{2-6} 
 &
  FaceDs~\cite{jourabloo2018face} &
  GAN,   Spoof trace, Depth &
  1.2 &
  1.7 &
  1.5 \\ \cline{2-6} 
 &
  STASN~\cite{yang2019face} &
  Region   Selection Module (RAM), CNN, LSTM &
  1.2 &
  2.5 &
  1.9 \\ \cline{2-6} 
 &
  MobileNet + Attention~\cite{chen2019attention} &
  Attention   based  feature fusuion, multi-scale   retinex (MSR) &
  3.9 &
  9.5 &
  6.7 \\ \cline{2-6} 
 &
  ResNet +Attention~\cite{chen2019attention} &
  Attention   based  feature fusuion, multi-scale   retinex (MSR) &
  5.1 &
  6.7 &
  5.9 \\ \cline{2-6} 
 &
  DeepPixBi~\cite{george2019deep} &
  Dense   Block, Binary mask &
  0.83 &
  0 &
  0.42 \\ \cline{2-6} 
 &
  BASN~\cite{kim2019basn} &
  Depth,   Reflection &
  {\color[HTML]{231F20} 1.5} &
  {\color[HTML]{231F20} 5.8} &
  {\color[HTML]{231F20} 3.6} \\ \cline{2-6} 
 &
  FAS-SGTD~\cite{wang2020deep} &
  Depth,   ConvGRU, Short-term Spatio-Temporal Block &
  2 &
  0 &
  1 \\ \cline{2-6} 
 &
  SpoofTrace~\cite{liu2020disentangling} &
  GAN,   Spoof trace &
  0.8 &
  1.3 &
  1.1 \\ \cline{2-6} 
 &
  Disentangled~\cite{zhang2020face} &
  Generative   models: Disentangled representation, Depth, LBP, Reconstruction loss &
  1.7 &
  0.8 &
  1.3 \\ \cline{2-6} 
 &
  BCN~\cite{yu2020face} &
  Depth,   Reflection, Bilateral Convolutional Networks (BCN), Multi-level feature   refinement &
  0 &
  1.6 &
  0.8 \\ \cline{2-6} 
 &
  DFST~\cite{huang2020deep} &
  Depth,   Frequency response &
  2.2 &
  0 &
  1.1 \\ \cline{2-6} 
 &
  SAPLC~\cite{sun2020face} &
  Background   separation, Binary mask &
  0 &
  0.83 &
  0.42 \\ \cline{2-6} 
 &
  CDCN~\cite{yu2020searching} &
  Depth,   Central difference convolution (CDC), Multi scale attention module (MAFM) &
  0.4 &
  1.7 &
  1 \\ \cline{2-6} 
 &
  CDCN++~\cite{yu2020searching} &
  Depth,   CDC, MAFM, Neural architectual search (NAS) &
  {\color[HTML]{231F20} 0.4} &
  {\color[HTML]{231F20} 0} &
  {\color[HTML]{231F20} \textbf{0.2}} \\ \cline{2-6} 
 &
  NAS-FAS~\cite{yu2020fas} &
  Central   difference pooling (CDP), CDC, Static-Dynamic image, MAFM, NAS, Depth &
  0.4 &
  0 &
  \textbf{0.2} \\ \cline{2-6} 
 &
  OCA-FAS~\cite{qin2020one} &
  Meta-learner,   binary mask &
  1.5 &
  1.3 &
  1.4 \\ \cline{2-6} 
 &
  CDCN-PS~\cite{yu2021revisiting} &
  CDCN,   Pyramid supervision, Binary mask &
  0.4 &
  1.2 &
  0.8 \\ \cline{2-6} 
\multirow{-19}{*}{1} &
  S-CNN+PL+TC w/ 50 samples~\cite{quan2021progressive} &
  Progressive   learning (PL), Temporal constraint (TC), Adaptive transfer (AT) &
  0.6±0.4 &
  0.0±0.0 &
  0.4±0.2 \\ \hline
 &
  Auxiliary~\cite{liu2018learning} &
  Depth,   rPPg &
  2.7 &
  2.7 &
  2.7 \\ \cline{2-6} 
 &
  FaceDs~\cite{jourabloo2018face} &
  GAN,   Spoof trace, Depth &
  4.2 &
  4.4 &
  4.3 \\ \cline{2-6} 
 &
  STASN~\cite{yang2019face} &
  Region   Selection Module (RAM), CNN, LSTM &
  4.2 &
  0.3 &
  2.2 \\ \cline{2-6} 
 &
  MobileNet + Attention~\cite{chen2019attention} &
  Attention   based  feature fusuion, multi-scale   retinex (MSR) &
  3.6 &
  9 &
  6.3 \\ \cline{2-6} 
 &
  ResNet + Attention~\cite{chen2019attention} &
  Attention   based  feature fusuion, multi-scale   retinex (MSR) &
  7.6 &
  2.2 &
  4.9 \\ \cline{2-6} 
 &
  DeepPixBi~\cite{george2019deep} &
  Dense   Block, Binary mask &
  11.39 &
  0.56 &
  5.97 \\ \cline{2-6} 
 &
  BASN~\cite{kim2019basn} &
  Depth,   Reflection &
  2.4 &
  3.1 &
  2.7 \\ \cline{2-6} 
 &
  FAS-SGTD~\cite{wang2020deep} &
  Depth,   ConvGRU, Short-term Spatio-Temporal Block &
  2.5 &
  1.3 &
  1.9 \\ \cline{2-6} 
 &
  Disentangled~\cite{zhang2020face} &
  GAN,   Spoof trace &
  1.1 &
  3.6 &
  2.4 \\ \cline{2-6} 
 &
  SpoofTrace~\cite{liu2020disentangling} &
  Generative   models: Disentangled representation, Depth, LBP, Reconstruction loss &
  2.3 &
  1.6 &
  1.9 \\ \cline{2-6} 
 &
  BCN~\cite{yu2020face} &
  Depth,   Reflection, Bilateral Convolutional Networks (BCN), Multi-level feature   refinement &
  2.6 &
  0.8 &
  1.7 \\ \cline{2-6} 
 &
  DFST~\cite{huang2020deep} &
  Depth,   Frequency response &
  2.3 &
  1 &
  1.7 \\ \cline{2-6} 
 &
  SAPLC~\cite{sun2020face} &
  Background   separation, Binary mask &
  2.78 &
  2.22 &
  2.5 \\ \cline{2-6} 
 &
  CDCN~\cite{yu2020searching} &
  Depth,   Central difference convolution (CDC), Multi scale attention module (MAFM) &
  1.5 &
  1.4 &
  1.5 \\ \cline{2-6} 
 &
  CDCN++~\cite{yu2020searching} &
  Depth,   CDC, MAFM, Neural architectual search (NAS) &
  1.8 &
  0.8 &
  1.3 \\ \cline{2-6} 
 &
  NAS-FAS~\cite{yu2020fas} &
  Central   difference pooling (CDP), CDC, Static-Dynamic image, MAFM, NAS, Depth &
  1.5 &
  0.8 &
  \textbf{1.2} \\ \cline{2-6} 
 &
  OCA-FAS~\cite{qin2020one} &
  Meta-learner,   binary mask &
  - &
  - &
  - \\ \cline{2-6} 
 &
  CDCN-PS~\cite{yu2021revisiting} &
  CDCN,   Pyramid supervision, Binary mask &
  1.4 &
  1.4 &
  1.4 \\ \cline{2-6} 
\multirow{-19}{*}{2} &
  S-CNN+PL+TC  w/ 50 samples~\cite{quan2021progressive} &
  Progressive   learning (PL), Temporal constraint (TC), Adaptive transfer (AT) &
  1.7±0.9 &
  0.6±0.3 &
  1.2±0.5 \\ \hline
 &
  Auxiliary~\cite{liu2018learning} &
  Depth,   rPPg &
  2.7±1.3 &
  3.1±1.7 &
  2.9±1.5 \\ \cline{2-6} 
 &
  FaceDs~\cite{jourabloo2018face} &
  GAN,   Spoof trace, Depth &
  4.0±1.8 &
  3.8±1.2 &
  3.6±1.6 \\ \cline{2-6} 
 &
  STASN~\cite{yang2019face} &
  Region   Selection Module (RAM), CNN, LSTM &
  4.7±3.9 &
  0.9±1.2 &
  2.8±1.6 \\ \cline{2-6} 
 &
  MobileNet + Attention~\cite{chen2019attention} &
  Attention   based  feature fusuion, multi-scale   retinex (MSR) &
  8.7±4.5 &
  5.3±2.3 &
  6.3±2.2 \\ \cline{2-6} 
 &
  ResNet + Attention~\cite{chen2019attention} &
  Attention   based  feature fusuion, multi-scale   retinex (MSR) &
  3.9±2.8 &
  7.3±1.1 &
  5.6±1.6 \\ \cline{2-6} 
 &
  DeepPixBi~\cite{george2019deep} &
  Dense   Block, Binary mask &
  11.67±19.57 &
  10.56±14.1 &
  11.11±9.4 \\ \cline{2-6} 
 &
  BASN~\cite{kim2019basn} &
  Depth,   Reflection &
  {\color[HTML]{231F20} 1.8±1.1} &
  {\color[HTML]{231F20} 3.6±3.5} &
  {\color[HTML]{231F20} 2.7±1.6} \\ \cline{2-6} 
 &
  FAS-SGTD~\cite{wang2020deep} &
  Depth,   ConvGRU, Short-term Spatio-Temporal Block &
  3.2±2.0 &
  2.2±1.4 &
  2.7±0.6 \\ \cline{2-6} 
 &
  Disentangled~\cite{zhang2020face} &
  GAN,   Spoof trace &
  2.8±2.2 &
  1.7±2.6 &
  2.2±2.2 \\ \cline{2-6} 
 &
  SpoofTrace~\cite{liu2020disentangling} &
  Generative   models: Disentangled representation, Depth, LBP, Reconstruction loss &
  1.6±1.6 &
  4.0±5.4 &
  2.8±3.3 \\ \cline{2-6} 
 &
  BCN~\cite{yu2020face} &
  Depth,   Reflection, Bilateral Convolutional Networks (BCN), Multi-level feature   refinement &
  2.8±2.4 &
  2.3±2.8 &
  2.5±1.1 \\ \cline{2-6} 
 &
  DFST~\cite{huang2020deep} &
  Depth,   Frequency response &
  4.4±3.7 &
  2.0±1.4 &
  3.2±1.4 \\ \cline{2-6} 
 &
  SAPLC~\cite{sun2020face} &
  Background   separation, Binary mask &
  {\color[HTML]{231F20} 4.72±4.24} &
  {\color[HTML]{231F20} 3.06±3.47} &
  {\color[HTML]{231F20} 3.89±2.11} \\ \cline{2-6} 
 &
  CDCN~\cite{yu2020searching} &
  Depth,   Central difference convolution (CDC), Multi scale attention module (MAFM) &
  2.4±1.3 &
  2.2±2.0 &
  2.3±1.4 \\ \cline{2-6} 
 &
  CDCN++~\cite{yu2020searching} &
  Depth,   CDC, MAFM, Neural architectual search (NAS) &
  1.7±1.5 &
  2.0±1.2 &
  1.8±0.7 \\ \cline{2-6} 
 &
  NAS-FAS~\cite{yu2020fas} &
  Central   difference pooling (CDP), CDC, Static-Dynamic image, MAFM, NAS, Depth &
  2.1±1.3 &
  1.4±1.1 &
  \textbf{1.7±0.6} \\ \cline{2-6} 
 &
  OCA-FAS (Ours) &
  Meta-learner,   binary mask &
  1.5±1.4 &
  3.5±1.3 &
  2.5±1.4 \\ \cline{2-6} 
 &
  CDCN-PS~\cite{yu2021revisiting} &
  CDCN,   Pyramid supervision, Binary mask &
  1.9±1.7 &
  2.0±1.8 &
  2.0±1.7 \\ \cline{2-6} 
\multirow{-19}{*}{3} &
  S-CNN+PL+TC w/ 50 samples~\cite{quan2021progressive} &
  Progressive   learning (PL), Temporal constraint (TC), Adaptive transfer (AT) &
  1.5±0.9 &
  2.2±1.0 &
  1.7±0.8 \\ \hline
 &
  Auxiliary~\cite{liu2018learning} &
  Depth,   rPPg &
  9.3±5.6 &
  10.4±6.0 &
  9.5±6.0 \\ \cline{2-6} 
 &
  FaceDs ~\cite{jourabloo2018face} &
  GAN,   Spoof trace, Depth &
  1.2±6.3 &
  6.1±5.1 &
  5.6±5.7 \\ \cline{2-6} 
 &
  STASN~\cite{yang2019face} &
  Region   Selection Module (RAM), CNN, LSTM &
  6.7±10.6 &
  8.3±8.4 &
  7.5±4.7 \\ \cline{2-6} 
 &
  MobileNet + Attention~\cite{chen2019attention} &
  Attention   based  feature fusuion, multi-scale   retinex (MSR) &
  10.9±4.6 &
  12.7±5.1 &
  11.3±3.9 \\ \cline{2-6} 
 &
  ResNet + Attention~\cite{chen2019attention} &
  Attention   based  feature fusuion, multi-scale   retinex (MSR) &
  11.3±3.9 &
  9.7±4.8 &
  9.8±4.2 \\ \cline{2-6} 
 &
  DeepPixBi~\cite{george2019deep} &
  Dense   Block, Binary mask &
  36.67±29.67 &
  13.33±16.75 &
  25.0±12.67 \\ \cline{2-6} 
 &
  BASN~\cite{kim2019basn} &
  Depth,   Reflection &
  {\color[HTML]{231F20} 6.4±8.6} &
  {\color[HTML]{231F20} 3.2±5.3} &
  {\color[HTML]{231F20} 4.8±6.4} \\ \cline{2-6} 
 &
  FAS-SGTD~\cite{wang2020deep} &
  Depth,   ConvGRU, Short-term Spatio-Temporal Block &
  6.7±7.5 &
  3.3±4.1 &
  5.0±2.2 \\ \cline{2-6} 
 &
  Disentangled~\cite{zhang2020face} &
  GAN,   Spoof trace &
  5.4±2.9 &
  3.3±6.0 &
  4.4±3.0 \\ \cline{2-6} 
 &
  SpoofTrace~\cite{liu2020disentangling} &
  Generative   models: Disentangled representation, Depth, LBP, Reconstruction loss &
  2.3±3.6 &
  5.2±5.4 &
  3.8±4.2 \\ \cline{2-6} 
 &
  BCN~\cite{yu2020face} &
  Depth,   Reflection, Bilateral Convolutional Networks (BCN), Multi-level feature   refinement &
  2.9±4.0 &
  7.5±6.9 &
  5.2±3.7 \\ \cline{2-6} 
 &
  DFST~\cite{huang2020deep} &
  Depth,   Frequency response &
  9.3±3.7 &
  7.5±3.7 &
  8.4±2.7 \\ \cline{2-6} 
 &
  SAPLC~\cite{sun2020face} &
  Background   separation, Binary mask &
  {\color[HTML]{231F20} 11.94±6.93} &
  {\color[HTML]{231F20} 6.67±5.52} &
  {\color[HTML]{231F20} 9.31±4.42} \\ \cline{2-6} 
 &
  CDCN~\cite{yu2020searching} &
  Depth,   Central difference convolution (CDC), Multi scale attention module (MAFM) &
  4.6±4.6 &
  9.2±8.0 &
  6.9±2.9 \\ \cline{2-6} 
 &
  CDCN++~\cite{yu2020searching} &
  Depth,   CDC, MAFM, Neural architectual search (NAS) &
  4.2±3.4 &
  5.8±4.9 &
  5.0±2.9 \\ \cline{2-6} 
 &
  NAS-FAS~\cite{yu2020fas} &
  Central   difference pooling (CDP), CDC, Static-Dynamic image, MAFM, NAS, Depth &
  4.2±5.3 &
  1.7±2.6 &
  \textbf{2.9±2.8} \\ \cline{2-6} 
 &
  OCA-FAS~\cite{qin2020one} &
  Meta-learner,   binary mask &
  2.3±2.5 &
  5.9±4.5 &
  4.1±2.7 \\ \cline{2-6} 
 &
  CDCN-PS~\cite{yu2021revisiting} &
  CDCN,   Pyramid supervision, Binary mask &
  2.9±4.0 &
  5.8±4.9 &
  4.8±1.8 \\ \cline{2-6} 
\multirow{-19}{*}{4} &
  S-CNN+PL+TC w/ 50 samples~\cite{quan2021progressive} &
  Progressive   learning (PL), Temporal constraint (TC), Adaptive transfer (AT) &
  5.2±2.0 &
  4.6±4.1 &
  4.8±2.0 \\ \hline
\end{tabular}%
}
\end{table}

\begin{table}[H]
\centering
\caption{Comparison of face anti-spoofing methods on four testing sets for domain generalization on face anti-spoofing.}
\label{tab:inter-dataset1}
\resizebox{\textwidth}{!}{%
\begin{tabular}{|c|c|cc|cc|cc|cc|}
\hline
\multirow{2}{*}{Method} &
  \multirow{2}{*}{Cues} &
  \multicolumn{2}{c|}{O\&C\&I to M} &
  \multicolumn{2}{c|}{O\&M\&I to C} &
  \multicolumn{2}{c|}{O\&C\&M to I} &
  \multicolumn{2}{c|}{I\&C\&M to O} \\ \cline{3-10} 
 &
   &
  \multicolumn{1}{c|}{HTER(\%)} &
  AUC(\%) &
  \multicolumn{1}{c|}{HTER(\%)} &
  AUC(\%) &
  \multicolumn{1}{c|}{HTER(\%)} &
  AUC(\%) &
  \multicolumn{1}{c|}{HTER(\%)} &
  AUC(\% \\ \hline
Auxiliary   (Depth Only) ~\cite{liu2018learning} &
  Depth &
  \multicolumn{1}{c|}{22.72} &
  85.88 &
  \multicolumn{1}{c|}{33.52} &
  73.15 &
  \multicolumn{1}{c|}{29.14} &
  71.69 &
  \multicolumn{1}{c|}{30.17} &
  77.61 \\ \hline
Auxiliary   (All) ~\cite{liu2018learning} &
  Depth, rPPg &
  \multicolumn{1}{c|}{-} &
  - &
  \multicolumn{1}{c|}{28.4} &
  - &
  \multicolumn{1}{c|}{27.6} &
  - &
  \multicolumn{1}{c|}{-} &
  - \\ \hline
MADDG~\cite{shao2019multi} &
  \begin{tabular}[c]{@{}c@{}}Domain discriminator, Advarsarial learning,\\  Triplet loss, Depth\end{tabular} &
  \multicolumn{1}{c|}{17.69} &
  88.06 &
  \multicolumn{1}{c|}{24.5} &
  84.51 &
  \multicolumn{1}{c|}{22.19} &
  84.99 &
  \multicolumn{1}{c|}{27.98} &
  80.02 \\ \hline
PAD-GAN~\cite{wang2020unsupervised} &
  \begin{tabular}[c]{@{}c@{}}GAN, Advarsarial learning, Disentangled representation,\\  Multi-domain feature learning\end{tabular} &
  \multicolumn{1}{c|}{17.02} &
  90.1 &
  \multicolumn{1}{c|}{19.68} &
  87.43 &
  \multicolumn{1}{c|}{20.87} &
  86.72 &
  \multicolumn{1}{c|}{25.02} &
  81.02 \\ \hline
RFMetaFAS~\cite{shao2020regularized} &
  Meta-learner, Depth &
  \multicolumn{1}{c|}{13.89} &
  93.98 &
  \multicolumn{1}{c|}{20.27} &
  88.16 &
  \multicolumn{1}{c|}{17.3} &
  90.48 &
  \multicolumn{1}{c|}{16.45} &
  91.16 \\ \hline
CCDD~\cite{saha2020domain} &
  \begin{tabular}[c]{@{}c@{}}Gradiant Reversal Layer (GRL),\\  Class Conditional Domain Discriminator (CCDD), \\ Advarsarial learning, LSTM\end{tabular} &
  \multicolumn{1}{c|}{15.42} &
  91.13 &
  \multicolumn{1}{c|}{17.41} &
  90.12 &
  \multicolumn{1}{c|}{15.87} &
  91.47 &
  \multicolumn{1}{c|}{14.72} &
  93.08 \\ \hline
NAS-FAS~\cite{yu2020fas} &
  \begin{tabular}[c]{@{}c@{}}Central Difference Pooling (CDP),\\  Central Difference Convolution (CDC),\\   Static-Dynamic image, NAS, Depth,\\ Multiscale Attention Fusion Module (MAFM)\end{tabular} &
  \multicolumn{1}{c|}{19.53} &
  88.63 &
  \multicolumn{1}{c|}{16.54} &
  90.18 &
  \multicolumn{1}{c|}{14.51} &
  93.84 &
  \multicolumn{1}{c|}{13.8} &
  93.43 \\ \hline
NAS-FA /w   D-MetaS~\cite{yu2020fas} &
  NAS-FAS, Meta-Learner &
  \multicolumn{1}{c|}{16.85} &
  90.42 &
  \multicolumn{1}{c|}{15.21} &
  92.64 &
  \multicolumn{1}{c|}{\textbf{11.63}} &
  \textbf{96.98} &
  \multicolumn{1}{c|}{13.16} &
  94.18 \\ \hline
SSDG-R~\cite{jia2020single} &
  \begin{tabular}[c]{@{}c@{}}Gradiant Reversal Layer (GRL), Advarsarial learning,\\   Domain Discriminator, Triplet loss, L2 normalize\end{tabular} &
  \multicolumn{1}{c|}{\textbf{7.38}} &
  \textbf{97.17} &
  \multicolumn{1}{c|}{\textbf{10.44}} &
  \textbf{95.94} &
  \multicolumn{1}{c|}{11.71} &
  96.59 &
  \multicolumn{1}{c|}{15.61} &
  91.54 \\ \hline
SDA*~\cite{wang2021self} &
  Meta-learner,  Adaptor, Depth &
  \multicolumn{1}{c|}{15.4} &
  91.8 &
  \multicolumn{1}{c|}{24.5} &
  84.4 &
  \multicolumn{1}{c|}{15.6} &
  90.1 &
  \multicolumn{1}{c|}{23.1} &
  84.3 \\ \hline
DR-UDA*~\cite{wang2020unsupervised} &
  \begin{tabular}[c]{@{}c@{}}Distangled Representation, \\ Advarsarial learning, K-NN classifier\end{tabular} &
  \multicolumn{1}{c|}{16.1} &
  - &
  \multicolumn{1}{c|}{22.2} &
  - &
  \multicolumn{1}{c|}{22.7} &
  - &
  \multicolumn{1}{c|}{24.7} &
  - \\ \hline
DASN~\cite{kim2021suppressing} &
  \begin{tabular}[c]{@{}c@{}}Spoof-irrelavant  features, Gradiant Reversal Layer (GRL), \\ Discriminator heads, Advarsarial learning\end{tabular} &
  \multicolumn{1}{c|}{8.33} &
  96.31 &
  \multicolumn{1}{c|}{12.04} &
  95.33 &
  \multicolumn{1}{c|}{13.38} &
  86.63 &
  \multicolumn{1}{c|}{\textbf{11.77}} &
  \textbf{94.65} \\ \hline
S-CNN+PL+TC+AT*~\cite{quan2021progressive} &
  \begin{tabular}[c]{@{}c@{}}Progressive learning (PL), Temporal constraint (TC), \\  Adaptive transfer (AT)\end{tabular} &
  \multicolumn{1}{c|}{7.82±1.21} &
  \textbf{\underline{97.67±1.09}} &
  \multicolumn{1}{c|}{\textbf{\underline{4.01±0.81}}} &
  \textbf{\underline{98.96±0.77}} &
  \multicolumn{1}{c|}{\textbf{\underline{10.36±1.86}}} &
  \textbf{\underline{97.16±1.04}} &
  \multicolumn{1}{c|}{14.23±0.98} &
  93.66±0.75 \\ \hline
OCA-FAS~\cite{qin2020one} &
  Meta-learner, Binary mask &
  \multicolumn{1}{c|}{15.51} &
  91.92 &
  \multicolumn{1}{c|}{18.45} &
  89.33 &
  \multicolumn{1}{c|}{19.51} &
  88.29 &
  \multicolumn{1}{c|}{19.25} &
  88.55 \\ \hline
CDCN-PS~\cite{yu2021revisiting} &
  CDCN, Pyramid supervision, Binary mask &
  \multicolumn{1}{c|}{20.42} &
  87.43 &
  \multicolumn{1}{c|}{18.25} &
  86.76 &
  \multicolumn{1}{c|}{19.55} &
  86.38 &
  \multicolumn{1}{c|}{15.76} &
  92.43 \\ \hline
\end{tabular}%
}
{\tiny \raggedright * Models have access to target domain in unsupervised manner.\\ Underlined values are the best without differentiating * methods from other ones.  \par}
\end{table}

\begin{table}[H]
\centering
\caption{Cross testing on CASIA-MFSD vs. Replay-Attack based on HTER(\%).}
\label{tab:inter-dataset2}
\resizebox{\textwidth}{!}{%
\begin{tabular}{|c|c|c|cc|cc|}
\hline
 &
  {\color[HTML]{231F20} } &
  {\color[HTML]{231F20} } &
  \multicolumn{1}{c|}{{\color[HTML]{231F20} Train}} &
  {\color[HTML]{231F20} Test} &
  \multicolumn{1}{c|}{{\color[HTML]{231F20} Train}} &
  {\color[HTML]{231F20} Test} \\ \cline{4-7} 
\multirow{-2}{*}{Training data} &
  \multirow{-2}{*}{{\color[HTML]{231F20} Method}} &
  \multirow{-2}{*}{{\color[HTML]{231F20} Cues}} &
  \multicolumn{1}{c|}{{\color[HTML]{231F20} \begin{tabular}[c]{@{}c@{}}CASIA\\ MFSD\end{tabular}}} &
  {\color[HTML]{231F20} \begin{tabular}[c]{@{}c@{}}Replay\\  Attack\end{tabular}} &
  \multicolumn{1}{c|}{{\color[HTML]{231F20} \begin{tabular}[c]{@{}c@{}}Replay\\  Attack\end{tabular}}} &
  {\color[HTML]{231F20} \begin{tabular}[c]{@{}c@{}}CASIA\\ MFSD\end{tabular}} \\ \hline
 &
  {\color[HTML]{231F20} Auxiliary~\cite{liu2018learning}} &
  {\color[HTML]{231F20} Depth, rPPg} &
  \multicolumn{2}{c|}{27.60} &
  \multicolumn{2}{c|}{28.40} \\ \cline{2-7} 
 &
  {\color[HTML]{231F20} FaceDs~\cite{jourabloo2018face}} &
  {\color[HTML]{231F20} GAN, Spoof trace, Depth} &
  \multicolumn{2}{c|}{28.50} &
  \multicolumn{2}{c|}{41.10} \\ \cline{2-7} 
 &
  {\color[HTML]{231F20} STASN~\cite{yang2019face}} &
  {\color[HTML]{231F20} \begin{tabular}[c]{@{}c@{}}Region Selection Module (RAM),\\  CNN, LSTM\end{tabular}} &
  \multicolumn{2}{c|}{31.50} &
  \multicolumn{2}{c|}{30.90} \\ \cline{2-7} 
 &
  MobileNet + Attention~\cite{chen2019attention} &
  \begin{tabular}[c]{@{}c@{}}Attention-based feature fusuion, \\ multi-scale retinex (MSR)\end{tabular} &
  \multicolumn{2}{c|}{30.00} &
  \multicolumn{2}{c|}{33.40} \\ \cline{2-7} 
 &
  ResNet-18 +Attention~\cite{chen2019attention} &
  \begin{tabular}[c]{@{}c@{}}Attention-based feature fusuion, \\ multi-scale retinex (MSR)\end{tabular} &
  \multicolumn{2}{c|}{36.20} &
  \multicolumn{2}{c|}{34.70} \\ \cline{2-7} 
 &
  {\color[HTML]{231F20} BASN~\cite{kim2019basn}} &
  {\color[HTML]{231F20} Depth, Reflection} &
  \multicolumn{2}{c|}{23.60} &
  \multicolumn{2}{c|}{29.90} \\ \cline{2-7} 
 &
  {\color[HTML]{231F20} FAS-SGTD~\cite{wang2020deep}} &
  {\color[HTML]{231F20} \begin{tabular}[c]{@{}c@{}}Depth, ConvGRU, \\ Short-term Spatio-Temporal Block\end{tabular}} &
  \multicolumn{2}{c|}{17.00} &
  \multicolumn{2}{c|}{\textbf{22.80}} \\ \cline{2-7} 
 &
  {\color[HTML]{231F20} BCN~\cite{yu2020face}} &
  {\color[HTML]{231F20} \begin{tabular}[c]{@{}c@{}}Depth, Reflection, \\ Bilateral Convolutional Networks (BCN),\\ Multi-level feature refinement\end{tabular}} &
  \multicolumn{2}{c|}{16.60} &
  \multicolumn{2}{c|}{36.40} \\ \cline{2-7} 
 &
  {\color[HTML]{231F20} Disentangled~\cite{zhang2020face}} &
  {\color[HTML]{231F20} \begin{tabular}[c]{@{}c@{}}Disentangled representation, \\ Depth, LBP, Reconstruction loss\end{tabular}} &
  \multicolumn{2}{c|}{22.40} &
  \multicolumn{2}{c|}{30.30} \\ \cline{2-7} 
 &
  {\color[HTML]{231F20} SALPC~\cite{sun2020face}} &
  {\color[HTML]{231F20} Background separation, Binary mask} &
  \multicolumn{2}{c|}{27.31} &
  \multicolumn{2}{c|}{37.50} \\ \cline{2-7} 
 &
  {\color[HTML]{231F20} SfSNet~\cite{sun2020face}} &
  {\color[HTML]{231F20} Depth, Reflectance, Albedo} &
  \multicolumn{2}{c|}{9.80} &
  \multicolumn{2}{c|}{29.60} \\ \cline{2-7} 
 &
  {\color[HTML]{231F20} CDCN~\cite{yu2020searching}} &
  {\color[HTML]{231F20} \begin{tabular}[c]{@{}c@{}}Depth, Central difference convolution (CDC), \\ Multi-scale Attention Fusion Module (MAFM)\end{tabular}} &
  \multicolumn{2}{c|}{15.50} &
  \multicolumn{2}{c|}{32.60} \\ \cline{2-7} 
\multirow{-20}{*}{\begin{tabular}[c]{@{}c@{}}Solely trained\\  on source domain\end{tabular}} &
  {\color[HTML]{231F20} CDCN++~\cite{yu2020searching}} &
  {\color[HTML]{231F20} \begin{tabular}[c]{@{}c@{}}Depth, CDC, MAFM, \\ Neural Architectural Search (NAS)\end{tabular}} &
  \multicolumn{2}{c|}{\textbf{6.50}} &
  \multicolumn{2}{c|}{29.80} \\ \hline
 &
  {\color[HTML]{231F20} ML-MMD~\cite{zhou2019face}} &
  {\color[HTML]{231F20} Mutli-layer MMD} &
  \multicolumn{2}{c|}{33.10} &
  \multicolumn{2}{c|}{34.3} \\ \cline{2-7} 
 &
  Wang et al.~\cite{wang2019improving} &
  \begin{tabular}[c]{@{}c@{}}Domain discriminator,  Adversarial learning, \\ Multi-scale features, Triplet loss, K-NN classifier\end{tabular} &
  \multicolumn{2}{c|}{17.50} &
  \multicolumn{2}{c|}{41.60} \\ \cline{2-7} 
 &
  DR-UDA~\cite{wang2020unsupervised} &
  \begin{tabular}[c]{@{}c@{}}Domain Discriminator,  Adversarial learning, \\ Disentangled representation, Center loss, \\ Triplet loss, Reconstruction Loss, K-NN classifier\end{tabular} &
  \multicolumn{2}{c|}{15.60} &
  \multicolumn{2}{c|}{34.20} \\ \cline{2-7} 
 &
  {\color[HTML]{231F20} CCDD~\cite{saha2020domain}} &
  {\color[HTML]{231F20} \begin{tabular}[c]{@{}c@{}}Class Conditional Domain Discriminator (CCDD),\\  Gradiant Reversal Learning (GRL),\\    Adversarial Learning, LSTM\end{tabular}} &
  \multicolumn{2}{c|}{\textbf{11.38}} &
  \multicolumn{2}{c|}{\textbf{16.11}} \\ \cline{2-7} 
 &
  PAD-GAN~\cite{wang2020cross} &
  \begin{tabular}[c]{@{}c@{}}GAN, Advarsarial  learning, \\ Disentangeld representation, Reconstruction loss\end{tabular} &
  \multicolumn{2}{c|}{26.10} &
  \multicolumn{2}{c|}{39.20} \\ \cline{2-7} 
 &
  USDAN-Un~\cite{jia2021unified} &
  \begin{tabular}[c]{@{}c@{}}Domain discriminator, Gradiant Reversal Layer (GRL),\\  Advarsarial learning, Semanic  Alignment loss,\\  l2 normalize, Adaptive cross-entropy\end{tabular} &
  \multicolumn{2}{c|}{16.00} &
  \multicolumn{2}{c|}{30.20} \\ \cline{2-7} 
\multirow{-13}{*}{\begin{tabular}[c]{@{}c@{}}Domain adaptation \\ using target data\end{tabular}} &
  USDAN-Se (3-shot)*~\cite{jia2021unified} &
  \begin{tabular}[c]{@{}c@{}}Domain discriminator, Gradiant Reversal Layer (GRL),\\  Adversarial learning, Semantic Alignment loss,\\    l2 normalize, Adaptive cross-entropy\end{tabular} &
  \multicolumn{2}{c|}{\textbf{\underline{6.50}}} &
  \multicolumn{2}{c|}{\textbf{\underline{7.20}}} \\ \hline
\end{tabular}%
}
{\tiny \raggedright * The model has access to a few labeled data from target domain.\\ Underlined values are the best without differentiating * method from other DA methods.   \par}
\end{table}

\newpage
\section{Limitations and Possible Future Directions}\label{sec:limitations}
With the emergence of new face anti-spoofing techniques and numerous public datasets, researchers have shifted their focus from simple academic evaluations (intra-dataset evaluations on small datasets) to more challenging experiments (inter-dataset evaluations with large datasets). However, since a single erroneous decision in FAS systems will result in invalid access to an attacker, they must perform flawlessly. Therefore, in the following, we mention the existing obstacles that prevent FAS models from reaching industrial performance.
In other words, we try to provide some insights about some critical issues in FAS systems that can be considered as new research paths.

\paratitle{Datasets:} In this survey, we have mentioned different methods that have been employed to capture liveness-related features (e.g. Auxiliary Tasks) or to find a generalized feature space (e.g. Domain generalization and adaptation). Nevertheless, the models' performance is measured based on a limited number of datasets that cannot represent the diversity in real-world applications. As a result, despite acceptable performance in inter-dataset evaluations of some methods (Table \ref{tab:inter-dataset1}), they can fail to remain generalized when encountered with samples with significant variations in race, sensor, PIE, etc; thus, accurate evaluation of FAS methods requires extensive datasets. Although numerous FAS datasets exist(majority covered in section~\ref{sec:datasets}),

On one hand, numerous existing FAS datasets (majority of them covered in section~\ref{sec:datasets}) lack the diversity and quantity, necessary for comprehensive evaluation of FAS models. On the other hand, existing large-scale datasets~\cite{zhang2020celeba, patel2016secure} are in image format and consequently, they wont be useful for evaluation of video-based methods. Extensive datasets in video format not only can be employed for more accurate benchmarking, but may also be leveraged for training.

\paratitle{Model confidence}:
Deep learning approaches have enabled FAS systems to achieve desirable performance where the softmax function in the classification layer squeezes the final score between 0 and 1. This value is sometimes regraded as decision probability that is problematic especially when the model is over-confident for its errors. In such cases that samples might belong to an unrelated data distribution, it would be beneficial if the model could estimate the prediction uncertainty. 
A rational solution would be Bayesian models~\cites{van2021bayesian} that can mathematically reason about model's confidence.
A good approximation for Bayesian models can be Monte Carlo dropout~\cite{gal2016dropout}. In this technique, the drop out, used as a regularization  technique in training phase, is also enabled for inference and the final decision is based on the output of \emph{T} stochastic forward passes through the network and the average of results. 

Classification Uncertainty Quantification~\cites{sensoy2018evidential} as another solution for uncertainty modeling, is widely used in various real-world applications such as autonomous vehicles and is applied to both classification~\cites{sensoy2018evidential} and regression tasks~\cite{amini2019deep}.
However the model's confidence provided by these approaches, is crucial in context of FAS, it has never been discussed or considered in retrospective works. FAS systems usually focus on the accuracy, ignoring the potential advantages brought by decision uncertainty. These advantages include but are not limited to:
\begin{itemize}
\item Confidence value can be used as a measure to models' reliability for its predictions.
\item Low-confidence live predictions can be considered as PAs at the cost of higher FRR.
\item Low-confidence predictions of a specific method in ensemble methods can be discarded not to degrade the whole model's accuracy.
\end{itemize}

\paratitle{Domain adaptation:} Domain adaptation methods demonstrate promising performance in our inter-dataset evaluations (Tables~\ref{tab:inter-dataset1} and \ref{tab:inter-dataset2}), but they suffer from following drawbacks:
\begin{itemize}
\item Multiple adaptations: Each time domain shift occurs, models need a chance to adapt to new data distribution. While the adaptation process may not be plausible in some applications, the adaptation to multiple target domains can also become troublesome. If we consider a case that a model first adapts itself to domain \textit{A}, and then adapts itself to domain \textit{B}, there is no guarantee that the final feature space is shared among all target domains and source domains. Even if we merge and consider both target domains as one, the adaptation on the target domain can become cumbersome over time.
\item Dependence on source domain: domain adaptation methods attempt to close the gap between source and target domain, but as we can see in inter-dataset evaluation in table~\ref{tab:inter-dataset2}, domain adaptation methods' performance significantly vary when source and target domain are interchanged. For instance, in the inter-dataset test between CASIA-MFSD and Replay Attack, the methods have a dramatic performance drop when it is trained on Replay Attack unless the model has access to a few labeled data from target domain (semi-supervised case in~\cite{jia2021unified}), which may not be feasible in many scenarios.
\end{itemize}

To address the first mentioned issue, the same strategy as Perez et al.~\cites{perez2020learning} can be applied in which they leverage a buffer in their continual learning framework to preserve information from previous samples when trained on new samples. Thus, selected representative samples from previous target domains can be used for training when model is adapted on a new target domain. To cover the second problem, Jia et al.\cite{jia2021unified} have demonstrated having few labeled samples from target domain can be very helpful. On the other hand, Quan et al.\cite{quan2021progressive} proposed a transfer learning method to assign pseudo-label to target domain. Therefore, if a model have reliable confidence for its prediction, samples with high confidence in target domain can be used in supervised manner to resolve the second issue.

\textbf{Temporal information:} As we can see in evaluation results (Table \ref{tab:oulu},~\ref{tab:inter-dataset1}, and~\ref{tab:inter-dataset2}), a few methods utilize temporal cues in their proposed frameworks~\cite{yang2019face, wang2020deep, yu2020fas, huang2020deep}, and other existing works suffice to spatial information. Exploiting temporal information may increase models’ robustness, especially against static attacks, but it requires multiple frames or a video of a user. Thus, it results into increased processing time, which is not ideal for real-time applications. Therefore, to keep temporal information without too much computations, researchers should focus on more optimized implementations or low-computation solution. For instance, yu et al.\cites{yu2020searching} introduced static-dynamic image, which encapsulates temporal information from other frames in one frame. Therefore, the computational cost in the network is as same as single frame, with minimum preprocessing to acquire static-dynamic image. 

\textbf{Feature Extractor:} Most recent methods leverage auxiliary information (depth) or loss functions (adversarial loss, triplet loss, etc.) to help the feature generator to capture domain-independent liveness features. Although manipulating feature space have proven to be helpful, enriching quality of features can further improve general performance. However, majority of deep-learning-based studies, use vanilla convolution to extract features, and only few recent approaches have made an effort to optimize convolutional features, In~\cite{yu2020searching} and~\cite{yu2020fas} Central Difference Convolution and Central Difference Pooling was introduced, respectively. Since convolution and pooling operations are at the core of CNN networks, any optimization on them could result in significant improvements. Therefore, likewise low-level optimizations of basic building blocks of CNN deserves more attention.

\section{Conclusion}\label{sec:conclusion}
In this survey, we provided a comprehensive review of the most recent works in FAS where we focused on recent deep learning methods. Besides, we gave a detailed description of most major public datasets. To shed light to retrospective research efforts, we demonstrated different methods' performance based on the most challenging intra-dataset and inter-dataset benchmarks. Finally, we pointed out the limitations in existing approaches and datasets and suggested possible directions for further studies.

\recalctypearea

\printbibliography

\end{document}